\documentclass[lettersize,journal]{IEEEtran}
\usepackage{amsmath,amsfonts}
\usepackage{algpseudocode}
\usepackage{algorithm}
\usepackage{array}
\usepackage{subcaption}
\usepackage{textcomp}
\usepackage{stfloats}
\usepackage{verbatim}
\usepackage{graphicx}
\usepackage{cite}
\usepackage{hyperref}
\usepackage{adjustbox} 
\usepackage{booktabs}
\usepackage{multirow}
\usepackage[table,xcdraw]{xcolor}
\usepackage{xcolor}
\usepackage{pifont}
\usepackage{cite}
\usepackage{tikz}
\usetikzlibrary{shapes.multipart}
\usetikzlibrary{shapes,arrows}
\usepackage{amsmath}
\usepackage{bm}

\begin{document}

\title{CalibRefine: Deep Learning-Based Online Automatic Targetless LiDAR–Camera Calibration with Iterative and Attention-Driven Post-Refinement}

\author{Lei Cheng, Lihao Guo, Tianya Zhang, Tam Bang, Austin Harris, Mustafa Hajij, Mina Sartipi,~\IEEEmembership{Senior Member,~IEEE}, and Siyang Cao,~\IEEEmembership{Senior Member,~IEEE}
\thanks{Lei Cheng, Lihao Guo, and Siyang Cao are with the Department of Electrical and Computer Engineering, The University of Arizona, Tucson, AZ 85721 USA (e-mail: leicheng@arizona.edu; leolihao@arizona.edu; caos@arizona.edu)}
\thanks{Tianya Zhang, Tam Bang, Austin Harris, and Mina Sartipi are with the Center For Urban Informatics and Progress (CUIP), UTC Research Institute, University of Tennessee at Chattanooga, TN 37405 USA (e-mail: tianya-zhang@utc.edu; nlg643@mocs.utc.edu; austin-p-harris@utc.edu; mina-sartipi@utc.edu)}
\thanks{Mustafa Hajij is with Electrical Engineering Department, University of San Francisco, San Francisco, CA, 94117 USA (e-mail: mhajij@usfca.edu)}
}

\markboth{Journal of \LaTeX\ Class Files,~Vol.X, No.X, X}%
{Shell \MakeLowercase{\textit{et al.}}: A Sample Article Using IEEEtran.cls for IEEE Journals}


\maketitle

\begin{abstract}
Accurate multi-sensor calibration is essential for deploying robust perception systems in applications such as autonomous driving and intelligent transportation. Existing LiDAR-camera calibration methods often rely on manually placed targets, preliminary parameter estimates, or intensive data preprocessing, limiting their scalability and adaptability in real-world settings. In this work, we propose a fully automatic, targetless, and online calibration framework, \emph{CalibRefine}, which directly processes raw LiDAR point clouds and camera images. Our approach is divided into four stages: (1) a Common Feature Discriminator that leverages relative spatial positions, visual appearance embeddings, and semantic class cues to identify and generate reliable LiDAR-camera correspondences, (2) a coarse homography-based calibration that uses the matched feature correspondences to estimate an initial transformation between the LiDAR and camera frames, serving as the foundation for further refinement, (3) an iterative refinement to incrementally improve alignment as additional data frames become available, and (4) an attention-based refinement that addresses non-planar distortions by leveraging a Vision Transformer and cross-attention mechanisms. Extensive experiments on two urban traffic datasets demonstrate that CalibRefine achieves high-precision calibration with minimal human input, outperforming state-of-the-art targetless methods and matching or surpassing manually tuned baselines. Our results show that robust object-level feature matching, combined with iterative refinement and self-supervised attention-based refinement, enables reliable sensor alignment in complex real-world conditions without ground-truth matrices or elaborate preprocessing. Code is available at 
\href{https://github.com/radar-lab/Lidar\_Camera\_Automatic\_Calibration}{https://github.com/radar-lab/Lidar\_Camera\_Automatic\_Calibration}

\end{abstract}


\section{Introduction}
\IEEEPARstart{R}{eliable} and accurate environment perception is crucial for applications such as autonomous driving, robotics, and intelligent transportation systems, enabling informed decisions and ensuring safe, efficient operations. However, single-sensor perception often encounters inherent limitations \cite{zhou2024targetless,lv2021lccnet,sengupta2022robust}. Cameras provide rich visual detail but are sensitive to lighting variations and struggle with depth estimation, especially under poor illumination conditions \cite{berrio2021camera,zhang2023automated}. LiDAR sensors, conversely, provide precise 3D geometric measurements robust to lighting changes \cite{gong2023scene}, yet can be costly and suffer performance degradation under adverse weather. Consequently, multi-sensor fusion, particularly LiDAR-camera fusion, has gained prominence, integrating visual textures with accurate spatial data \cite{cheng2024deep,liu2022targetless}. Nonetheless, successful fusion fundamentally relies on accurate sensor calibration, as imprecise calibration severely compromises downstream perception accuracy \cite{zhao2024extrinsic,yoon2021targetless}.

Calibration procedures generally include intrinsic, extrinsic, and temporal calibration \cite{pervsic2021spatiotemporal}. While intrinsic calibration (determining internal sensor parameters) and temporal calibration (synchronizing timestamps) typically follow standardized practices and achieve reliable results \cite{lv2022observability,yuan2022licas3,yeong2021sensor}, extrinsic calibration—also known as spatial calibration—remains challenging. Extrinsic calibration seeks to identify spatial transformations between sensor coordinate systems, usually by establishing correspondences between matched points \cite{cheng20233d,yang2024autonomous}.
Extrinsic calibration methods vary primarily by how these correspondences are established: target-based approaches utilize dedicated calibration artifacts (e.g., checkerboards or markers) for precise correspondences, yet
they require elaborate setup and are impractical for real-world scenarios \cite{xiao2024calibformer,cheng2023online}; targetless methods instead use
natural scene features, eliminating cumbersome preparations but posing challenges when suitable features are scarce or indistinct \cite{cheng2023online,chen2022pbacalib}. Methods also differ regarding human intervention (manual versus automatic) and operational mode (offline versus online). Manual methods, although precise, involve significant human effort and thus lack scalability \cite{yin2023automatic}. Automatic methods autonomously establish feature correspondences, minimizing manual intervention, making them attractive for scalable and continuous operation  \cite{sun2022atop,zhang20232}. Offline calibration methods rely on batch data and extensive optimizations but fail to adapt to real-time sensor shifts, environmental changes, or hardware reconfigurations. In contrast, online calibration continuously updates calibration parameters as new data arrives, accommodating dynamic conditions, albeit with increased computational complexity \cite{schneider2017regnet,zhu2023calibdepth,zhu2023robust}.

Given these trade-offs, an automatic, targetless, and online calibration paradigm combines the most desirable attributes—removing calibration objects, eliminating human intervention, and adapting to changing environments. However, despite its appeal, such a paradigm remains highly challenging, and existing approaches in this category exhibit various limitations. Motion-based methods often rely on additional hardware or specific sensor motion constraints, limiting real-world applicability \cite{sun2022automatic}. Hand-eye calibration, while classical, demands multiple accurate sensor poses, rendering it unsuitable for static or dynamically constrained setups \cite{liu2024transformer}. Edge-based methods suffer from unreliable edge matching across
modalities, since object boundaries differ significantly across LiDAR and camera data \cite{li2024edgecalib,zhu2025targetless,yang2024autonomous}. Mutual information-based methods are similarly unreliable due to varying LiDAR reflectance and camera illumination sensitivities \cite{pandey2012automatic,koide2023general}. Recent deep learning–based methods directly regress calibration parameters (e.g., RegNet \cite{schneider2017regnet} and its variants \cite{zhu2023calibdepth,lv2021lccnet,xiao2024calibformer,10328060,8593693,9956145}) but require initial manual calibrations or projected depth maps, and suffer from limited generalization and computational overhead, making them less suitable for real-time applications \cite{xiao2024calibformer}.

\begin{figure*}[]
	\centering
	\includegraphics[width=\textwidth]{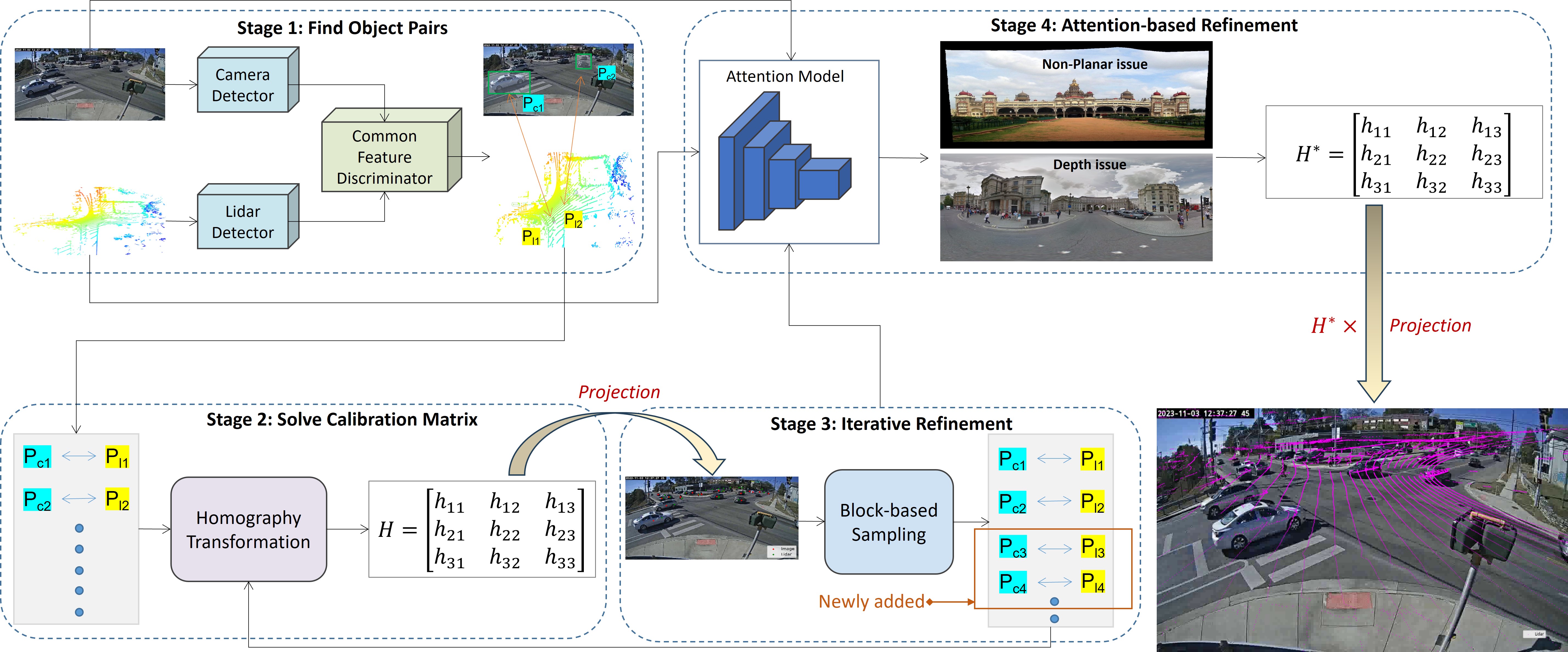}
	\caption{Work-Flow of the Proposed CalibRefine Framework for Fully Automatic Online Targetless LiDAR-Camera Calibration.}
	\label{framework}
\end{figure*}

\begin{figure*}[]
    \centering
    \begin{subfigure}[t]{0.53\textwidth}
        \centering
        \includegraphics[width=\textwidth]{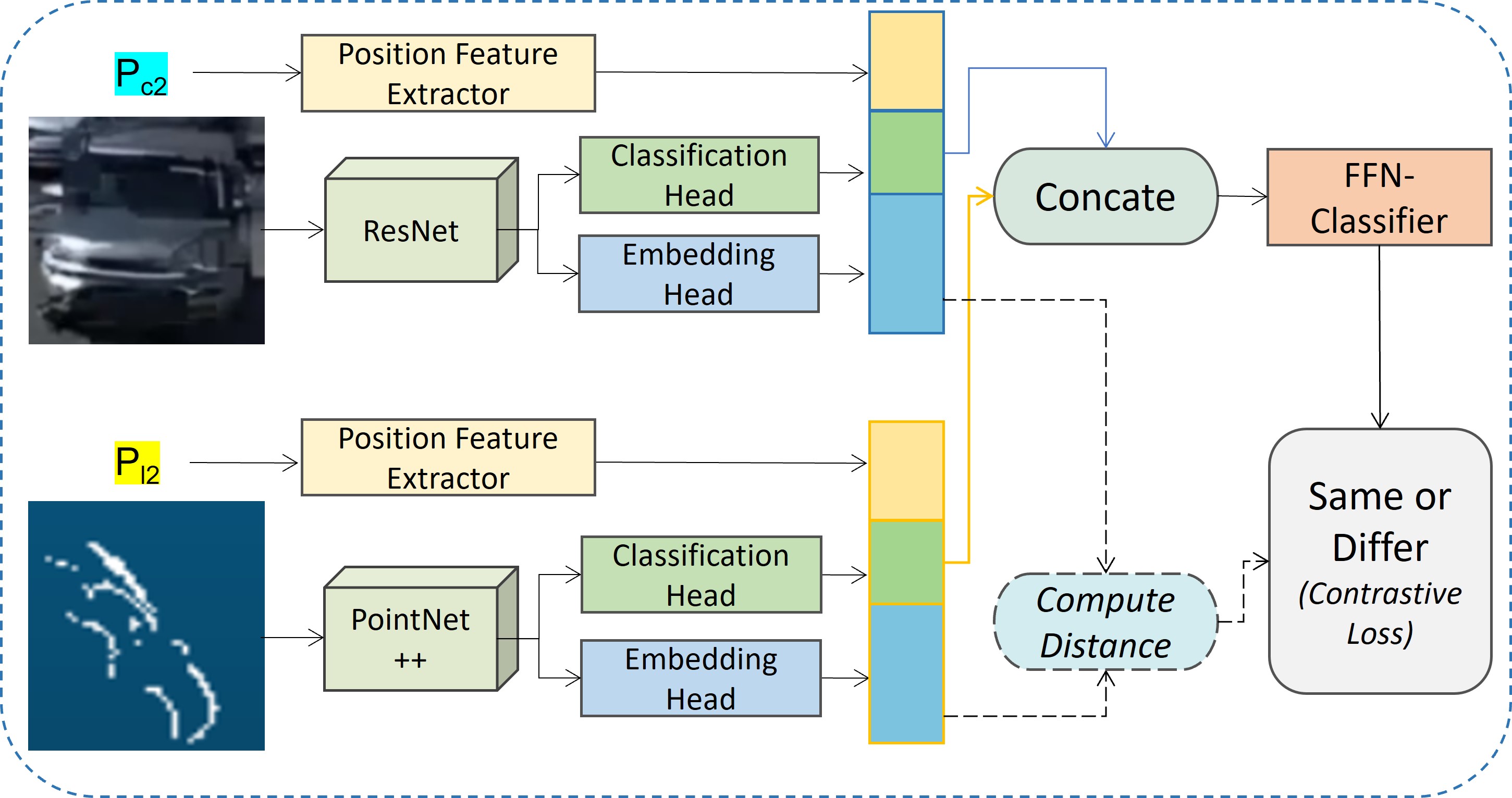}
        \caption{}
        \label{fig10:1}
    \end{subfigure}\hspace{0.001\textwidth}
    \begin{subfigure}[t]{0.46\textwidth}
        \centering
        \includegraphics[width=\textwidth]{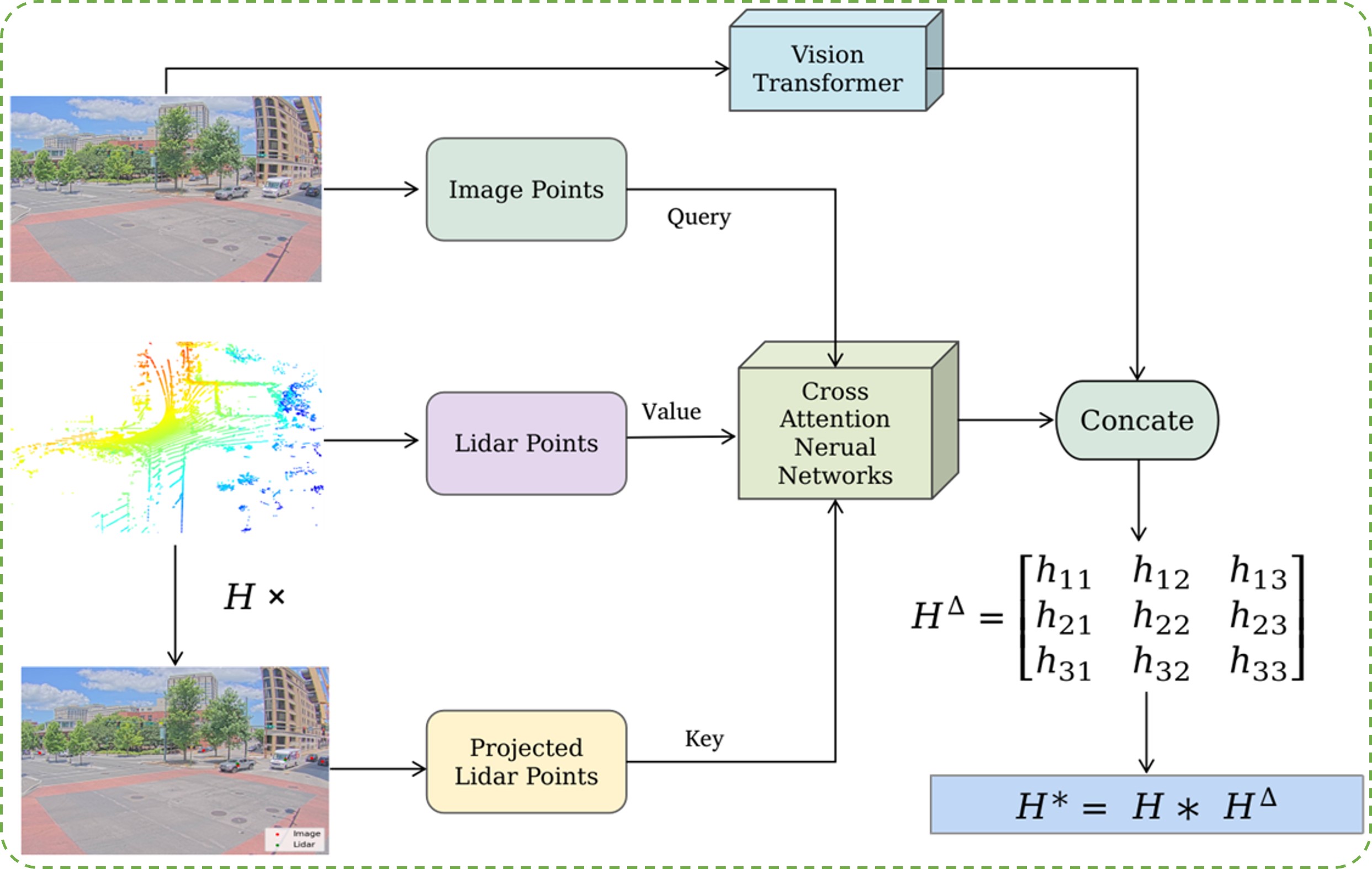}
        \caption{}
        \label{fig10:2}
    \end{subfigure}
    \caption{Key Components of Our Proposed Method: (a) Common Feature Discriminator, (b) Attention-based Refinement Module. }
    \label{fig:model_plot}
\end{figure*}

To address these challenges, we propose \emph{CalibRefine}, a fully automatic, targetless, online LiDAR–camera calibration framework, as shown in Fig.~\ref{framework} and \ref{fig:model_plot}. Our method directly processes raw LiDAR point clouds and camera images without initial calibration matrices or complex preprocessing. First, we leverage established object detection algorithms—YOLOv8 for camera data and an octree-based DBSCAN approach for LiDAR—to identify individual objects. A novel Common Feature Discriminator (CFD) then matches these cross-sensor object instances by learning relative positions, appearance embeddings (using ResNet~\cite{he2015deepresiduallearningimage} for camera images and PointNet++~\cite{qi2017pointnetdeephierarchicalfeature} for LiDAR data), and semantic class information, forming reliable cross-sensor correspondences.
Recognizing potential inaccuracies from initial matching, we further enhance calibration through two online refinement stages. The iterative refinement incrementally optimizes calibration using accumulated correspondence data, while the attention-based refinement employs a Vision Transformer~\cite{dosovitskiy2021imageworth16x16words} and cross-attention mechanisms to correct for non-planar distortions and depth variations, further improving calibration accuracy.
Crucially, our framework bypasses the pitfalls of direct matrix regression and the need for projected LiDAR maps, and eliminates the reliance on heuristic preprocessing or manually labeled calibration matrices, offering a more principled, data-driven pipeline that is both computationally efficient and adaptable in real-time. By integrating domain-specific mature object detection methods, a reliable discriminator to identify cross-sensor correspondences, and dual-stage refinement, our approach bridges the existing research gap, achieving a stable and accurate LiDAR–camera calibration that is truly automatic, targetless, and online. Our contributions are summarized as follows:  
\begin{enumerate}
\item \textbf{Fully Automatic, Targetless, and Online Calibration Framework}:  
   We propose a novel calibration framework that directly processes raw LiDAR point clouds and camera images, eliminating the need for heuristic preprocessing, manually labeled calibration matrices, or prior calibration. This ensures generalizability and adaptability across diverse scenarios.  

\item \textbf{Common Feature Discriminator for Accurate Cross-Sensor Matching}:  
   Our method introduces a deep learning–based Common Feature Discriminator to robustly identify shared object features across sensors by leveraging relative positions, appearance embeddings, and classification information, enabling precise object correspondences even in real-world environments.  

\item \textbf{Coarse-to-Fine Calibration Strategy with Dual Refinement Processes}: 
   The framework adopts a two-stage calibration approach, combining a homography-based coarse calibration with iterative refinement and attention-based refinement methods. These processes improve calibration accuracy in real-time, addressing challenges such as dynamic changes, densely distributed correspondences, and non-planar surfaces.  
  
\end{enumerate}

The remainder of this paper is organized as follows: Section 2 reviews the related works. Section 3 presents the proposed method in detail. Section 4 discusses the experimental results and analysis. Finally, Section 5 concludes the paper and outlines potential avenues for future research.

\section{Related Works}
Extrinsic calibration methods for LiDAR–camera systems can generally be divided into two categories: target-based and targetless approaches. Target-based calibration \cite{fu2019lidar,dong2024automatic} relies on specially designed calibration targets and is thus commonly associated with manual, offline procedures. Moreover, these methods typically cannot handle real-time decalibrations, which are common in practical scenarios. In contrast, targetless methods extract features directly from natural scenes, making them well-suited for automatic, online calibration.
These methods can be broadly categorized into motion-based, edge alignment-based, mutual information-based, and deep learning–based approaches.

\subsubsection{Motion-based Calibration}
Motion-based calibration methods leverage sensor movements or relative poses derived from visual and LiDAR odometry to compute extrinsic parameters~\cite{petek2024automatic,zhang2023overlap,yoon2021targetless,liu2024transformer}. For instance, Petek et al.~\cite{petek2024automatic} utilize odometry paths from each sensor, aligning them through non-linear optimization and dense 2D–3D matching. Zhang et al.~\cite{zhang2023overlap} similarly derive the calibration from relative sensor SLAM trajectories. Despite effectiveness under certain conditions, these methods heavily rely on accurate odometry or SLAM estimations, which are susceptible to noise and degenerate in scenarios with limited sensor motion (e.g., minimal rotation). Hand-eye calibration–based methods~\cite{liu2024transformer} further exacerbate this issue, requiring multiple precise sensor poses, thus limiting their practical applicability, particularly for static installations.

\subsubsection{Edge Alignment-based Calibration}
Edge-based approaches attempt to align edges detected from LiDAR point clouds and camera images~\cite{yang2024autonomous,li2024edgecalib,zhu2025targetless}. For example, Zhu et al.~\cite{zhu2025targetless} extract edge
features of LiDAR-Camera data for edge alignment based solely on laser reflectivity, while Li et al.~\cite{li2024edgecalib} employ advanced edge extraction techniques such as the Segment Anything Model (SAM) combined with multi-frame filtering. However, reliably matching edges between sensors is inherently challenging due to modality differences—LiDAR captures sparse geometric structures, whereas camera edges reflect dense texture and lighting variations. These differences often lead to inaccurate feature correspondence and suboptimal calibration outcomes.

\subsubsection{Mutual Information-based Calibration}
Mutual information (MI)-based calibration methods utilize statistical relationships between LiDAR reflectance and camera image intensities~\cite{pandey2012automatic,taylor2013automatic}. Pandey et al.~\cite{pandey2012automatic}, for instance, maximize mutual information between LiDAR reflectance intensities and camera images to find optimal alignment. Despite their conceptual elegance, these methods struggle due to significant variability in LiDAR reflectance caused by surface material differences and camera pixel intensity fluctuations arising from lighting changes, leading to inconsistent and less reliable calibration results.

\subsubsection{Deep Learning-based Calibration}
Deep learning-based methods have introduced neural networks to regress extrinsic calibration parameters directly~\cite{schneider2017regnet,lv2021lccnet,xiao2024calibformer,zhu2023robust,zhu2023calibdepth}. Seminal works such as RegNet~\cite{schneider2017regnet} and LCCNet~\cite{lv2021lccnet} regress calibration offsets using projected LiDAR depth maps derived from initial calibration estimates. Similarly, Xiao et al.~\cite{xiao2024calibformer} utilize transformer-based architectures to refine feature correspondences. Zhu et al.~\cite{zhu2023calibdepth} propose CalibDepth, which uses depth maps as a unified representation across modalities, integrating monocular depth estimation and sequence modeling to improve online calibration performance. However, these methods depend heavily on an initial calibration to project LiDAR point clouds into the image plane, forming a "projected LiDAR depth map" that aligns sparse LiDAR data with dense image pixels. While this enables cross-modal feature correlation, it significantly limits generalizability—since the initial calibration is often manually provided or empirically estimated. Moreover, due to the sparsity of LiDAR data, the resulting depth maps are dominated by image features, effectively sidelining useful LiDAR-specific information and heavily biases calibration toward image modality. In addition, direct regression tasks pose significant computational challenges due to their unconstrained nature, further limiting real-time applicability. Alternative semantic segmentation-based methods~\cite{luo2024zero} also face computational inefficiencies without substantial accuracy benefits over simpler object detection-based methods.

In summary, existing automatic, targetless, and online calibration methods commonly exhibit limitations including reliance on accurate odometry or sensor movement, difficulty in cross-modal edge matching, sensitivity to reflectance and illumination conditions, and computational inefficiencies. Furthermore,  most existing methods fail to fully exploit the advances in object detection and feature extraction developed individually for LiDAR and camera data processing.
To overcome these limitations, we propose \emph{CalibRefine}, a fully automatic, targetless, and online calibration framework that directly processes raw LiDAR point clouds and camera images, eliminating initial calibrations or elaborate preprocessing. Our approach integrates proven object detection algorithms and introduces a novel Common Feature Discriminator for robust cross-sensor correspondence matching. Furthermore, we employ a coarse-to-fine strategy combining iterative optimization and attention-driven refinement, enabling accurate and robust real-time calibration. By directly matching corresponding points across modalities, our approach facilitates a straightforward, one-shot, and end-to-end calibration process between the LiDAR and camera, significantly enhancing adaptability to real-world scenarios.

\section{Proposed Method}
\subsection{Problem Formulation}
\label{homo_sec}
Extrinsic calibration between sensors aims to unify detections from two different sensors into the same frame of reference or coordinate system, enabling the fusion of their respective detection information. 
LiDAR–camera extrinsic calibration is typically accomplished by solving for a transformation matrix that associates a point in the image pixel coordinate system (PCS) with its corresponding point in the LiDAR coordinate system (LCS). Since points in the LCS are 3D, while those in the PCS are 2D, most existing calibration methods rely on 3D-to-2D perspective projection. However, this approach has notable drawbacks.  
First, it requires the camera's intrinsic matrix, adding the burden of intrinsic camera calibration, which is often performed manually, thus hindering fully automatic extrinsic calibration. Second, estimating the 3D-to-2D transformation matrix is computationally more complex and prone to instability. More importantly, for most practical applications, 3D-to-2D perspective calibration is unnecessary for achieving effective LiDAR–camera data fusion. A simpler 2D-to-2D projective calibration, where the 2D LiDAR plane is obtained by removing the Z-axis, is sufficient.

\begin{figure}[]
	\centering
	\includegraphics[width=0.495\textwidth]{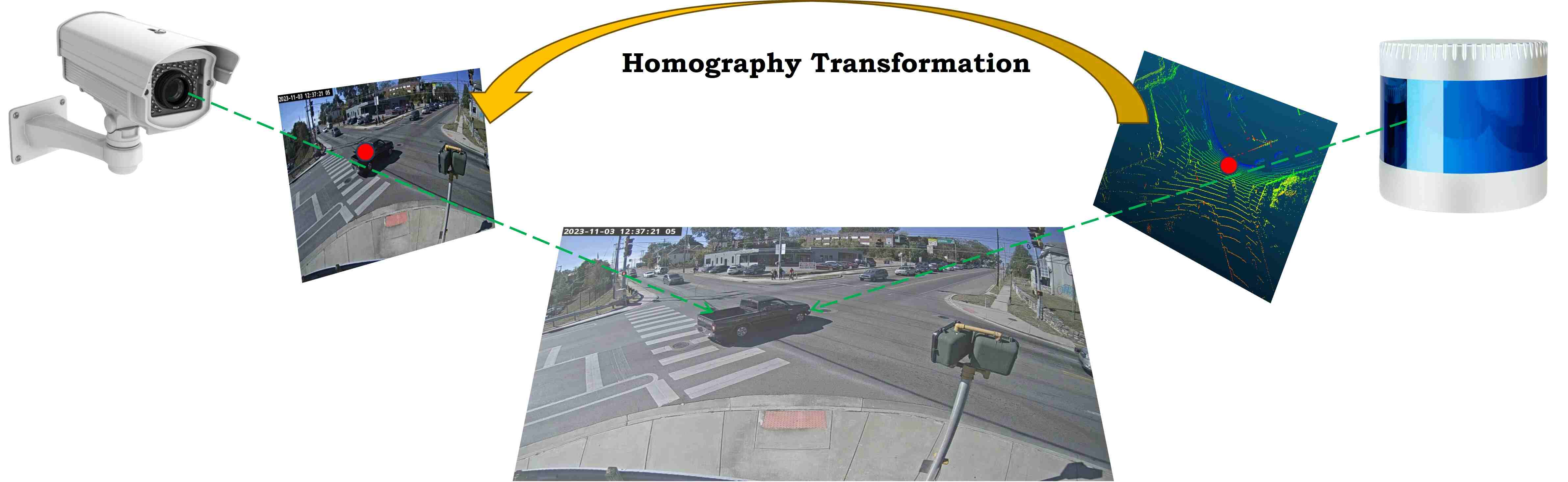}
	\caption{Illustration of Homography Transformation.}
	\label{HOMO}
\end{figure}

This simplification is justified for several reasons. The primary goal of calibration is to enable data fusion between the two sensors, such as associating 3D LiDAR point cloud clusters with image pixel regions for the same object. Achieving this does not require projecting the 3D LiDAR point cloud onto the image plane using a 3D-to-2D calibration matrix. Most existing methods adopt the 3D-to-2D approach as it draws from camera calibration practices that focus on 3D reconstruction. However, LiDAR–camera calibration is fundamentally different, as its focus is on data fusion, not reconstruction. By using 2D-to-2D projective calibration, where 2D LiDAR points are mapped to the image plane, corresponding 3D LiDAR points can still be retrieved without requiring a 3D-to-2D perspective transformation.  
Additionally, when projecting 3D LiDAR points onto the image plane, the LiDAR data effectively becomes 2D, resulting in the loss of LiDAR's inherent 3D detection capabilities. Therefore, 3D-to-2D calibration does not offer additional benefits over 2D-to-2D projective calibration. Notably, while many existing methods emphasize projecting 3D LiDAR points onto the image plane, this should only serve as a visualization tool to intuitively present calibration performance, not as the calibration objective itself. The true goal of calibration should be the seamless and accurate fusion of sensor data.

Thus, we propose using planar projective transformation to achieve 2D-to-2D calibration between the 2D LiDAR plane and the camera image plane, as illustrated in Fig. \ref{HOMO}. A planar projective transformation, or Homography, is an invertible linear transformation represented by a non-singular matrix \( \mathbf{H} \in \mathbb{R}^{3 \times 3} \) \cite{dubrofsky2009homography}. This transformation allows us to project a point in the LCS directly onto the camera image plane without requiring the camera intrinsic matrix. The relationship is expressed as:

\begin{equation}
\label{eq_H}
\begin{bmatrix}
\hat{u} \\
\hat{v} \\
1 
\end{bmatrix} 
= \mathbf{H}
\begin{bmatrix}
x \\
y \\
1 
\end{bmatrix}
= 
\begin{bmatrix}
h_{11} & h_{12} & h_{13} \\
h_{21} & h_{22} & h_{23} \\
h_{31} & h_{32} & h_{33}
\end{bmatrix}
\begin{bmatrix}
x \\
y \\
1 
\end{bmatrix},
\end{equation}
where (\( \hat{P_l} = (\hat{u},\hat{v}) \)) is the projection of a point \( P_l = (x, y) \) in the LCS onto the camera image plane PCS.
Notably, the objects or points on the 2D LiDAR plane and those on the camera image plane are derived from objects or points lying on a common plane (e.g., the ground plane), as shown in Fig. \ref{HOMO}. This alignment justifies the use of 2D Homography for LiDAR–camera calibration, as it can be considered a planar homography induced by the common plane \cite{szeliski2022computer}.
To solve for the Homography matrix, a set of \( N \) points in the LCS and their corresponding points in the PCS is required. 
Although 4 points are theoretically sufficient, using more points allows optimization of the solution via a cost function that minimizes the geometric reprojection error \cite{ye2021keypoint,dubrofsky2009homography}. This error, which measures the alignment between \( N \) projected LiDAR points and their corresponding 2D image pixel points (\( P_p = (u, v) \)), can be quantified either as the Average Euclidean Distance (AED):
\begin{equation}\label{eq:avg_error}
\begin{split}
\mathcal{E}_{\text{AED}} &= \frac{1}{N} \sum_{i=1}^{N} \left\| P_p^i - \hat{P}_l^i \right\|_{2} \\
&= \frac{1}{N} \sum_{i=1}^{N} \sqrt{(u^{i} - \hat{u}^{i})^2 + (v^{i} - \hat{v}^{i})^2},
\end{split}
\end{equation}
or as the Root Mean Square Error (RMSE):
\begin{equation}\label{eq:rmse_error}
\begin{split}
\mathcal{E}_{\text{RMSE}} &= \sqrt{\frac{1}{N} \sum_{i=1}^{N} \left\| P_p^i - \hat{P}_l^i \right\|_{2}^2} \\
&= \sqrt{\frac{1}{N} \sum_{i=1}^{N} \left[ (u^{i} - \hat{u}^{i})^2 + (v^{i} - \hat{u}^{i})^2 \right]}.
\end{split}
\end{equation}



\subsection{Method Overview}
We aim to develop a framework for LiDAR–camera online automatic targetless calibration that reduces human intervention, streamlines sensor integration, and ensures high precision in LiDAR–camera fusion applications. Our method comprises the following stages (see Fig.~\ref{framework} and \ref{fig:model_plot}): In \textbf{Stage~1}, the established LiDAR and camera detectors are utilized to detect and extract objects from their respective sensor data. Each detected object is extracted, including its center position (the bounding box center for the camera detection and the cluster center for the LiDAR point cloud). These objects are then used to train a Common Feature Discriminator, which determines whether an image object and a LiDAR object correspond to the same entity. To achieve this, the discriminator learns and compares three distinct features: Relative Positions, Appearance Embeddings, and Classification Information. These features are concatenated and passed through a feed-forward neural network (FFN) classifier, which outputs a decision on whether the objects from the two sensors are the same or different. \textbf{Stage~2} involves solving the calibration matrix using the identified object pairs. A homography transformation is applied to generate a coarse initial calibration matrix (\(H\)), which establishes a preliminary correspondence between objects detected by the LiDAR and camera. In \textbf{Stage~3}, an iterative refinement-based fine calibration is performed to refine the calibration matrix, considering that the Common Feature Discriminator may not precisely match all corresponding objects. The initial calibration matrix from the previous stage is used to project LiDAR data onto the camera plane, enabling the selection and construction of additional point correspondences based on distance criteria. These newly established correspondences are then used to achieve more precise calibration through iterative refinement. Finally, in \textbf{Stage~4}, attention-based refinement employs a Vision Transformer to extract global distortion features from images, addressing the limitations of homography calibration caused by non-planar surfaces and depth variations. This also compensates for errors introduced by the absence of camera intrinsic matrix-based rectification. Furthermore, it integrates a cross-attention network to compute weighted interactions between image pixels (as queries), projected LiDAR points (as keys), and LiDAR points (as values), thereby capturing more accurate correspondences between LiDAR and camera data points. The model fundamentally learns and generates a correction matrix (\(H^ \Delta\)) to refine the initial calibration, resulting in an improved matrix (\(H^*\)) for better LiDAR–camera alignment.

\subsection{Common Feature Discriminator}
The key to solving the extrinsic calibration matrix, which aligns the LiDAR and camera coordinate systems, lies in identifying a sufficient number of object correspondences between the two sensor views. Although objects detected by LiDAR and cameras may appear quite different due to the disparate nature of the data (geometric point clouds versus pixel-based images), they inherently share some common characteristics:  
\begin{enumerate}
    \item \textbf{Shape:} Objects exhibit geometric shapes that can be captured as contours in camera images and point clusters in LiDAR data.  
    \item \textbf{Semantic Information:} Both LiDAR and camera data can reveal high-level semantic features, such as object categories (e.g., vehicles, pedestrians), that correspond across modalities.  
    \item \textbf{Reflection Intensity:} LiDAR measures reflection intensity based on surface properties, while cameras capture similar information through brightness and contrast.  
\end{enumerate}
Recognizing and leveraging these shared features offers a viable approach to establishing robust correspondences \cite{cheng2025transrad} between LiDAR and camera detections of the same objects.

To achieve this, we propose the Common Feature Discriminator (see Fig. \ref{fig10:1}), a deep learning–based model that leverages advanced feature extraction to learn and extract shared features from LiDAR and camera data, thereby enabling effective object matching and correspondence identification. The first step is to detect and crop individual objects from each sensor’s output. For camera-based object detection, we adopt YOLOv8~\cite{Yolov8} to robustly detect objects in images and generate bounding boxes around them. In parallel, LiDAR-based object detection is performed using an octree-based change detection algorithm~\cite{octree} followed by DBSCAN clustering, which segments the point cloud into clusters, each hypothesized to belong to a distinct object. Since the LiDAR and camera frames are time-synchronized, each LiDAR cluster and corresponding camera bounding box at the same timestamp can be treated as candidate detections from complementary modalities.

Once the objects are cropped from both sensor outputs, they are fed into the Common Feature Discriminator, whose task is to determine whether an object in a camera image and an object in a LiDAR point cloud correspond to the same physical entity. To this end, the discriminator learns and compares three key types of features: relative positions, appearance embeddings, and classification information. In the LiDAR branch, each 3D point cluster \(\mathbf{X}_L\) is processed by a LiDAR backbone (e.g., PointNet++) that encodes its local and global geometric structure into a latent vector. A classification head then outputs the object category while an embedding head produces a 128-dimensional feature, yielding
\[
\mathbf{z}_L = f_{\mathrm{emb}}(\mathbf{X}_L), \quad \hat{c}_L = f_{\mathrm{cls}}(\mathbf{X}_L),
\]
where \(\mathbf{z}_L \in \mathbb{R}^{128}\) represents the LiDAR embedding and \(\hat{c}_L\) the predicted class.

Simultaneously, each camera-cropped object (the pixels within its bounding box) is processed by an image backbone (e.g., ResNet), which outputs both a 128-dimensional appearance embedding and a classification result:
\[
\mathbf{z}_C = g_{\mathrm{emb}}(\mathbf{I}_C), \quad \hat{c}_C = g_{\mathrm{cls}}(\mathbf{I}_C),
\]
where \(\mathbf{z}_C \in \mathbb{R}^{128}\) is the camera embedding and \(\hat{c}_C\) the predicted semantic class. In addition, both LiDAR and camera objects are passed through a position feature extractor that computes the relative positions from their respective 2D centers, \((x, y)\) for LiDAR and \((u, v)\) for the camera, resulting in a relative position vector
\[
\Delta \mathbf{p} = (u - x,\, v - y).
\]

These three types of features—relative positions, appearance embeddings, and classification information—are then concatenated into a unified feature vector:
\[
\mathbf{f} \;=\; \bigl[\Delta \mathbf{p};\, \mathbf{z}_L;\, \mathbf{z}_C;\, \hat{c}_L;\, \hat{c}_C\bigr],
\]
which is fed into a small feed-forward network (FFN) for binary classification (``Same'' vs.\ ``Differ''):
\[
\hat{o} \;=\; \sigma\!\bigl(\mathrm{FFN}(\mathbf{f})\bigr),
\]
where \(\sigma\) denotes the sigmoid activation. During training, a contrastive loss \(\mathcal{L}_{\mathrm{ctr}}\) encourages the embeddings of true matching pairs (\(\hat{o}=1\)) to be close, while pushing non-matching pairs (\(\hat{o}=0\)) apart.

By jointly analyzing \(\Delta \mathbf{p}\), \(\mathbf{z}_L\), \(\mathbf{z}_C\), and \(\hat{c}_L, \hat{c}_C\), the Common Feature Discriminator robustly determines whether the LiDAR and camera detections refer to the same underlying object, even when the modalities present substantially different raw representations, thereby enabling the system to automatically match and associate objects across LiDAR and camera views. This module, integrated with LiDAR and camera object detectors, constitutes the foundation of an {end-to-end} cross-sensor object matching workflow. Specifically, time-synchronized LiDAR and camera frames are processed in parallel, and bounding boxes or point clusters are cropped and fed into the discriminator to obtain pairwise correspondence labels. The resulting high-confidence matches form the cornerstone for computing the extrinsic calibration matrix that aligns the LiDAR and camera coordinate frames.


\begin{figure*}[]
	\centering
	\includegraphics[width=0.95\textwidth]{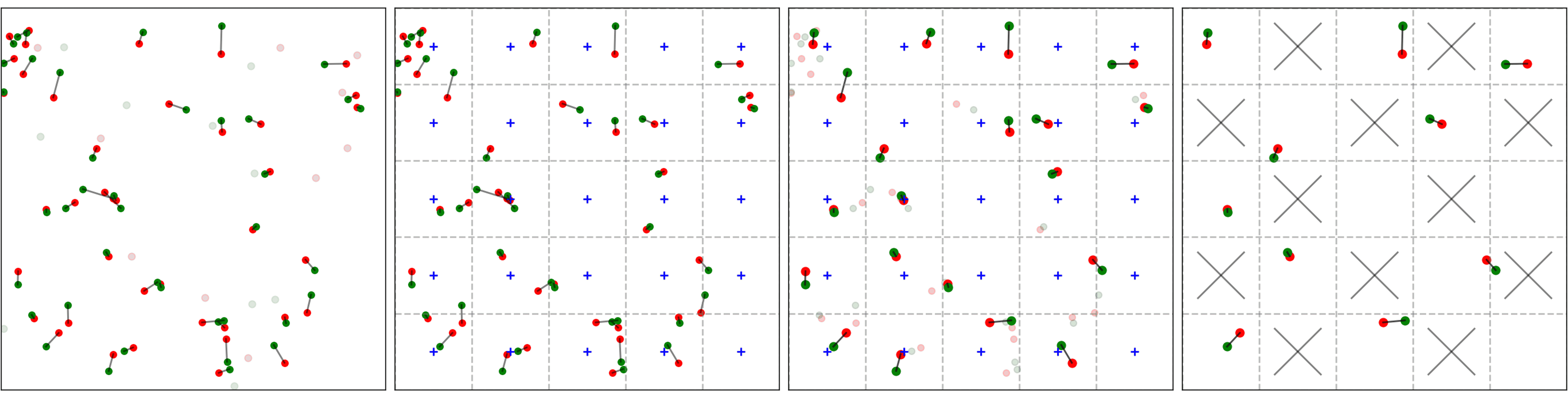}
	\caption{Block-based Sampling Strategy: 1) Project LiDAR points onto the image, identifying LiDAR-camera point pairs (red: camera, green: LiDAR); 2) Divide the image into equal-sized grids, marking centers; 3) Retain pairs whose camera point is nearest to the grid center; 4) Sample pairs at intervals of one block, discarding those in skipped blocks.}
	\label{BlockSampling}
\end{figure*}

\subsection{Homography-based Calibration Matrix Estimation}
\label{homo_est}
Once the Common Feature Discriminator identifies matched objects in the LiDAR and camera views, we extract their 2D center coordinates in each sensor’s frame to form point correspondences. Let us denote these correspondences by the set 
\[
\mathcal{C} 
= \Bigl\{\bigl(x_i,\,y_i\bigr) \leftrightarrow \bigl(u_i,\,v_i\bigr)\Bigr\}_{i=1}^{N},
\]
where \(\bigl(x_i,\,y_i\bigr)\) represents the \(i\)th LiDAR object center in the 2D LiDAR plane, and \(\bigl(u_i,\,v_i\bigr)\) denotes the corresponding camera object center in the image plane. Given these correspondences, we estimate the 2D homography matrix \(\mathbf{H}\in\mathbb{R}^{3\times 3}\) (cf.\ Eq.~\eqref{eq_H}) that satisfies
\[
\begin{bmatrix}
u_i \\[4pt]
v_i \\[2pt]
1
\end{bmatrix}
\;\approx\;
\mathbf{H}
\begin{bmatrix}
x_i \\[4pt]
y_i \\[2pt]
1
\end{bmatrix},
\quad
\text{for } i=1,\dots,N.
\]
To ensure robustness against erroneous matches, we employ the RANSAC algorithm \cite{cheng2023online} to iteratively fit \(\mathbf{H}\) while discarding outlier correspondences. Specifically, RANSAC randomly samples a small subset \(\mathcal{C}_s\subset\mathcal{C}\) of correspondences to compute a candidate \(\mathbf{H}_s\). It then evaluates \(\mathbf{H}_s\) on the entire set \(\mathcal{C}\) by measuring the reprojection error (e.g.\ \(\mathcal{E}_{\text{AED}}\) or \(\mathcal{E}_{\text{RMSE}}\)), and repeats this process over multiple iterations. The matrix \(\mathbf{H}\) yielding the largest inlier consensus (and thus the lowest average error) is ultimately selected.

Although RANSAC mitigates outliers, clustering of correspondences can still bias the homography solution if most matches lie in a small image region. To ensure that the point correspondences used in calibration are well-distributed across the sensor field of view—thus making the calibration results more representative and robust—we employ a \textit{block-based sampling} approach. As illustrated in Fig.~\ref{BlockSampling}, the camera image plane is partitioned into an array of blocks, each of size \(\delta_x\times \delta_y\) (\(5 \times 5\) in our case). Let
\[
\Omega = \bigcup_{j=1}^{J} B_j
\]
be the partition, where \(B_j\) is the \(j\)th block. For each block \(B_j\), we collect any point pairs whose camera coordinates \(\bigl(u_i,\,v_i\bigr)\) fall inside \(B_j\), then select exactly one representative \(\bigl(x_j^*,\,y_j^*\bigr)\leftrightarrow\bigl(u_j^*,\,v_j^*\bigr)\) nearest to \(B_j\)’s center \(\mathbf{c}_j\). This yields a spatially diverse subset
\[
\mathcal{C}' = \Bigl\{\bigl(x_j^*,y_j^*\bigr)\leftrightarrow\bigl(u_j^*,v_j^*\bigr)\Bigr\}_{j=1}^{J},
\]
which contributes to a more robust and stable homography estimate.

By combining object-level correspondences \(\mathcal{C}\) (or \(\mathcal{C}'\)) and outlier rejection (RANSAC) with the block-based sampling, we obtain a reliable homography-based calibration matrix 
\(\mathbf{H}_{\text{coarse}}\). Notably, this coarse calibration method requires no manual intervention, enabling real-time online calibration that can effectively handle runtime decalibration. By integrating the Common Feature Discriminator with this homography-based approach, we achieve a fully automated calibration pipeline, which serves as an initial coarse calibration step.

It is worth emphasizing that, unlike many existing LiDAR–camera calibration methods that attempt to utilize every LiDAR point, our approach relies solely on the centers of detected objects. We adopt this strategy for two main reasons (also as explained in Section~\ref{homo_sec}). First, since the goal of calibration is to align LiDAR objects with camera objects, using object center points is already sufficient for establishing accurate correspondences; incorporating all LiDAR points does not provide any additional benefit for object association and can actually complicate the calibration matrix estimation process. Second, even though the calibration matrix is derived from object center points only, it can still be used to project the entire LiDAR point cloud onto the image plane. Moreover, this center-based approach naturally fits an object-level matching paradigm, especially considering that camera-detected objects lack corresponding point cloud data. By reducing the reliance on dense point sets and focusing on object centers, we gain more degrees of freedom to achieve a robust and flexible calibration outcome.

\renewcommand{\algorithmiccomment}[1]{\hfill$\triangleright$\textit{\textcolor{blue}{#1}}}

\begin{figure}[htbp]
    \centering
    \begin{minipage}{0.498\textwidth}
        \begin{algorithm}[H]
            \caption{Iterative Refined LiDAR--Camera Calibration}\label{alg:iter}
            \textbf{Input: } $\text{Frames} \, F = \{(L_i, C_i)\}$ for $i=1 \ldots T$ \Comment{LiDAR and camera points for $T$ frames}\\
            \quad\quad\;\; $H_0$ \Comment{Initial calibration matrix is equal to $H_{coarse}$}\\
            \quad\quad\;\; $N \in \mathbb{Z}$ \Comment{Number of frames to accumulate before recalibration}\\
            \quad\quad\;\; $Bsz$ \Comment{Block size for sampling strategy}\\
            \textbf{Output: } $H^*$ \Comment{Refined calibration matrix}

            \begin{algorithmic}[1]
                \State $H_{best} \gets H_0$ \Comment{Set initial matrix as the best}
                \State $\mathcal{A} \gets [\ ]$ \Comment{Accumulated set to store point pairs}

                \For{$i \in \{1, 2, \dots, T\}$} 
                    \State $P^L_i \gets \text{Project}(L_i, H_{best})$ 
                    \Comment{Project LiDAR points onto the camera plane}
                    
                    \State $\mathcal{M}_i \gets \text{GBMatch}(P^L_i, C_i)$ 
                    \Comment{Greedy matched point pairs}
                    
                    \State $\widetilde{\mathcal{M}}_i \gets \text{BBSample}(\mathcal{M}_i, Bsz)$ 
                    \Comment{Block-based sampling to filter point pairs}
                    
                    \State $\mathcal{A}.\text{append}(\widetilde{\mathcal{M}}_i)$ 
                    \Comment{Accumulate new pairs}
                    
                    \If {$i \bmod N == 0$}
                            \State $H_{new} \gets \text{Recalibrate}(\mathcal{A})$
                        \Comment{Recalibrate using accumulated pairs after N frames}
                        
                        \State $\mathcal{E}_{old} \gets \text{RepErr}(H_{best}, \mathcal{A})$ \Comment{Get reprojection error}
                        \State $\mathcal{E}_{new} \gets \text{RepErr}(H_{new}, \mathcal{A})$ \Comment{Get reprojection error}
                        
                        \If {$\mathcal{E}_{new} < \mathcal{E}_{old}$} 
                            \State $H_{best} \gets H_{new}$
                            \Comment{Update matrix if error reduces}
                        \EndIf
                    \EndIf
                \EndFor

                \State $H^* \gets H_{best}$
                \Comment{Final refined calibration matrix}
                \State \Return $H^*$

            \end{algorithmic}
        \end{algorithm}
    \end{minipage}
\end{figure}

\subsection{Iterative Refinement Process}
While relying on the Common Feature Discriminator to establish a coarse initial calibration matrix provides a strong starting point, it may not perfectly match every corresponding object between LiDAR and camera data. In practice, leveraging additional data points—thereby increasing redundancy and expanding field-of-view coverage—often improves both the accuracy and robustness of the calibration. To this end, we propose an iterative refinement procedure (as demonstrated in Algorithm \ref{alg:iter}) that successively updates the calibration matrix by incorporating newly discovered point correspondences across multiple frames.

We begin with the coarse calibration matrix, denoted as \(\mathbf{H}_0 = \mathbf{H}_{\mathrm{coarse}}\), and use the LiDAR–camera point pairs identified during the coarse calibration (by using the Common Feature Discriminator) to form an initial accumulated set \(\mathcal{A}\). For each incoming frame \((L_i, C_i)\), where \(i \in \{1, \dots, T\}\), every LiDAR object center \((x_j, y_j)\) in \(L_i\) is projected onto the camera plane using the current best calibration matrix \(\mathbf{H}_{\mathrm{best}}\) as follows:
\[
\begin{bmatrix}
\hat{u}_j \\[4pt]
\hat{v}_j \\[2pt]
1
\end{bmatrix}
\;=\;
\mathbf{H}_{\mathrm{best}}
\begin{bmatrix}
x_j \\[2pt]
y_j \\[2pt]
1
\end{bmatrix}.
\]
Each projected point \((\hat{u}_j, \hat{v}_j)\) is then compared with the camera-detected object centers in \(C_i\). A greedy bipartite matching algorithm \cite{besser2017greedy} is used to associate each projected LiDAR point with its nearest camera detection (if one exists) based on a distance measure \(d\bigl((\hat{u},\hat{v}),(u,v)\bigr)\). Let \(\mathcal{M}_i\) be the associated candidate point pairs set from frame \(i\). Only candidate point pairs that fall within unoccupied or sufficiently distinct grid regions—determined by our block-based sampling strategy (Fig.~\ref{BlockSampling})—are retained to form a filtered set \(\widetilde{\mathcal{M}}_i\subseteq \mathcal{M}_i\), which is then incorporated into the accumulated set \(\mathcal{A}\) via
\[
\mathcal{A} \leftarrow \mathcal{A} \cup \widetilde{\mathcal{M}}_i.
\]
After accumulating data from every \(N = 100\) frames (or another empirically chosen threshold), a new calibration matrix \(\mathbf{H}_{\mathrm{new}}\) is re-estimated from the set \(\mathcal{A}\) using the homography calibration algorithm described in Section~\ref{homo_est}:
\begin{align*}
\mathbf{H}_{\mathrm{new}} &= \mathrm{Recalibrate}\bigl(\mathcal{A}\bigr) \quad \text{via minimizing} \\
&\quad \sum_{(x, y) \leftrightarrow (u, v) \in \mathcal{A}} \varphi\bigl(\mathbf{H}, (x, y), (u, v)\bigr),
\end{align*}
where \(\varphi(\cdot)\) denotes the chosen reprojection error function (e.g., \(\mathcal{E}_{\mathrm{AED}}\) or \(\mathcal{E}_{\mathrm{RMSE}}\)). If \(\mathbf{H}_{\mathrm{new}}\) results in a reduced reprojection error, it replaces the current best matrix, i.e., \(\mathbf{H}_{\mathrm{best}} \leftarrow \mathbf{H}_{\mathrm{new}}\). This process is iterated for each subsequent frame until reaching the final time step \(T\), gradually refining the calibration matrix by incorporating newly validated point correspondences.

By systematically incorporating additional correspondences at each iteration, this optimization loop converges toward a more robust calibration matrix. It maintains the practical advantages of the initial deep learning–based matching while progressively enhancing accuracy through redundancy and extended spatial coverage. Moreover, its iterative nature naturally accommodates runtime changes in the environment, thus helping to mitigate potential decalibration over long-term operation. 
Notably, we opt for a greedy bipartite matching \cite{besser2017greedy} approach rather than the more popular {Hungarian algorithm} for several practical reasons. Ideally, each LiDAR detection would correspond to exactly one camera detection, and bipartite graph matching would produce a one-to-one mapping that minimizes the overall matching cost. However, real-world conditions deviate from this ideal scenario: variations in the field of view and detection capabilities can lead to certain objects being detected by only one sensor. For example, LiDAR may capture distant objects outside the camera’s range, whereas a camera may pick up small or reflective objects that the LiDAR cannot reliably detect.
Given these discrepancies, the goal of bipartite graph matching is to identify the best subset of matching pairs, without forcing all detections from both sensors to be paired. Greedy bipartite matching is well-suited to this task, as it prioritizes finding and accumulating the lowest-cost matches while allowing some objects to remain unmatched if no suitable pair exists. In contrast, the Hungarian algorithm aims for an {optimal, one-to-one, and complete} assignment—i.e., pairing every detection from both sensors—an assumption that does not hold in many real-world LiDAR–camera detections. Such forced one-to-one pairings can degrade matching quality when unmatchable objects are forced to pair with unrelated detections.


\subsection{Attention-based Refinement Process}
While homography-based calibration yields a reasonable initial solution, its reliance on planar assumptions often introduces significant errors in real-world environments characterized by complex depth variations. To overcome these limitations and further refine the calibration, we introduce an attention-based deep learning model that produces a correction matrix \(\mathbf{H}^\Delta\) (see Fig.~\ref{fig10:2}). The refined calibration is computed as 
\[
\mathbf{H}^* \;=\; \mathbf{H} \times \mathbf{H}^\Delta,
\]
where \(\mathbf{H}\) is the initial calibration matrix and \(\mathbf{H}^\Delta\) compensates for non-planar distortions, lens imperfections, and other real-world discrepancies.

Our approach leverages a Vision Transformer (ViT) to capture global distortion features. Given an image \(\mathbf{I}\) partitioned into \(m\) patches \(\{\mathbf{p}_1,\dots,\mathbf{p}_m\}\), each patch is encoded into a token \(\mathbf{t}_i=f_{\mathrm{ViT}}(\mathbf{p}_i)\). The ViT applies multi-head self-attention; for one head,
\[
\mathrm{Attn}(\mathbf{Q},\mathbf{K},\mathbf{V}) \;=\; \mathrm{Softmax}\!\Bigl(\frac{\mathbf{Q}\,\mathbf{K}^\top}{\sqrt{d}}\Bigr)\mathbf{V},
\]
where \(\mathbf{Q},\mathbf{K},\mathbf{V}\in\mathbb{R}^{m\times d}\) are the query, key, and value matrices obtained by learned linear projections of \(\{\mathbf{t}_i\}\), and \(d\) is the token feature dimension. This aggregates global information to reveal distortion patterns that planar models cannot capture. A global ViT representation is then obtained by average pooling,
\[
\mathbf{z}_{\mathrm{ViT}} \;=\; \frac{1}{m}\sum_{i=1}^{m}\mathbf{t}_i \in \mathbb{R}^{d}.
\]

Simultaneously, a cross-attention mechanism establishes precise correspondences between LiDAR and camera data. Image points \(\{(u_i,v_i)\}_{i=1}^{N}\) produce queries \(\mathbf{Q_c}\in\mathbb{R}^{N\times d}\); LiDAR points, projected to the image by the calibrated LiDAR-to-image matrix \(\mathbf{H}\), yield keys \(\mathbf{K_c}\in\mathbb{R}^{M\times d}\), and their 3D coordinates \(\{(x_j,y_j,z_j)\}_{j=1}^{M}\) provide values \(\mathbf{V_c}\in\mathbb{R}^{M\times d}\). The cross-attention output is
\[
\mathbf{A}_{\mathrm{cross}} \;=\; \mathrm{Softmax}\!\Bigl(\frac{\mathbf{Q_c}\,\mathbf{K_c}^\top}{\sqrt{d}}\Bigr)\mathbf{V_c},
\]
which links each image point to its corresponding LiDAR point.

The global ViT representation \(\mathbf{z}_{\mathrm{ViT}}\in\mathbb{R}^{d}\) and the cross-attention output \(\mathbf{A}_{\mathrm{cross}}\in\mathbb{R}^{N\times d}\) are concatenated to form a unified feature vector,
\[
\mathbf{f} \;=\; \bigl[\mathbf{z}_{\mathrm{ViT}},\, \mathbf{A}_{\mathrm{cross}}\bigr],
\]
which is then processed by a regression head \(g:\mathbb{R}^{2d}\!\to\!\mathbb{R}^{9}\) to regress a 9-dimensional vector \(\boldsymbol{\theta}\). This vector is reshaped to obtain the correction matrix,
\[
\boldsymbol{\theta} = g\!\bigl(\mathbf{f}\bigr), \ \
\mathbf{H}^\Delta \;=\; \mathrm{Reshape}\!\bigl(\boldsymbol{\theta}\bigr) \in \mathbb{R}^{3\times 3},
\]
resulting in the final refined homography \(\mathbf{H}^* = \mathbf{H} \times \mathbf{H}^\Delta\). During training, a self-supervised reprojection loss minimizes the discrepancy between \(\mathbf{H}^{*}\)-projected LiDAR points and their matched image coordinates. Let \(\tilde{\mathbf{u}}_{j}=\Pi(\mathbf{H}^{*};x_j,y_j,1)\in\mathbb{R}^{2}\) denote the homogeneous projection (with normalization) of the \(j\)-th LiDAR point and \(\mathbf{u}_{\pi(j)}=(u_{\pi(j)},v_{\pi(j)})\) its matched image point; we optimize
\[
\mathcal{L}_{\mathrm{reproj}} \;=\; \frac{1}{M}\sum_{j=1}^{M} \bigl\| \tilde{\mathbf{u}}_{j}-\mathbf{u}_{\pi(j)} \bigr\|_{2}^{2},
\]
which guides \(\mathbf{H}^{\Delta}\) to correct residual misalignment.

By combining global image-level context (from the Vision Transformer) with precise, point-level cross-attention, the model robustly captures spatial relationships in both 2D and 3D domains. This synergy accommodates complex depth variations and non-planar surfaces, corrects inaccuracies introduced by simpler homography assumptions, and increases resilience to real-world imaging conditions, such as partial occlusions or unrectified camera images without intrinsic parameters.
Another key advantage of our proposed attention-based deep learning model is that it can be trained in a {self-supervised} manner, without requiring explicit annotation of LiDAR–camera correspondences. Specifically, the model iteratively adjusts the homography matrix by comparing projected LiDAR points against their nearest image correspondences, allowing these {implicit} pairings to serve as the supervisory signal. Consequently, the model is able to autonomously learn a {correction matrix} \(H^{\Delta}\) that minimizes reprojection errors—i.e., discrepancies between the LiDAR points (projected into the camera frame) and their corresponding image points. By relying on these implicit constraints within the data itself—rather than manual annotations—our approach eliminates human effort and intervention, thus enabling real-time, online LiDAR–camera calibration. It is worth noting that a relatively accurate initial matrix is crucial for effective self-supervised training. Therefore, our attention-based refinement is strategically positioned after the iterative refinement process, ensuring a robust starting point for the training.

\section{Experimental Results and Analysis}
\subsection{Sensor Setup and Data Collection}
\begin{figure}[htbp]
	\centering
	\includegraphics[width=0.499\textwidth]{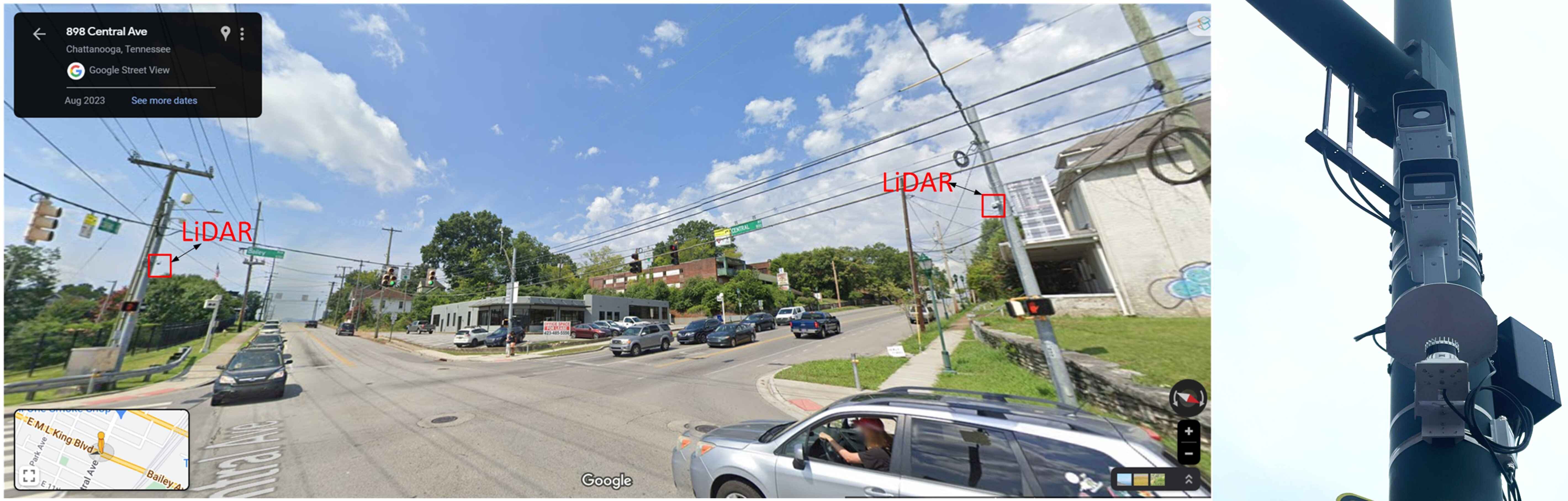}
	\caption{Sample Street-View of the Sensor Setup.} 
	\label{fig:sensorsetup}
\end{figure}
\textbf{Dataset 1} was collected at the intersection of M.L.K.\ Boulevard and Central Avenue in the Chattanooga Smart Corridor, where a two-hour synchronized dataset was gathered using multiple sensor types. A 32-channel LiDAR system was mounted on utility poles at the intersection corners (Fig.~\ref{fig:sensorsetup}), operating with a detection range of 0.05--120\,m and complemented by integrated video cameras. 

\textbf{Dataset 2} was collected at another urban intersection in downtown Chattanooga (Georgia Avenue and M.L.K.\ Boulevard), also employing a LiDAR--camera system. LiDAR scans and camera images were synchronized via ROS and stored in \textit{ROSbag} files with precise timestamps, ensuring consistent multi-modal alignment for cross-sensor calibration studies.

\subsection{Deep Learning Model Training}
\subsubsection{Data Annotation and Dataset Generation}
We developed a multi-sensor annotation toolkit for efficiently labeling common objects in both camera images and point cloud data. It combines automatic and manual annotation strategies to balance speed and labeling quality. In camera images, a YOLO-based algorithm automatically generates bounding boxes, which can be manually refined. For LiDAR data, background extraction and DBSCAN clustering detect objects, producing preliminary bounding boxes that are also subject to manual adjustment. Once detections from both sensors are complete, the toolkit provides a dual-view interface to match identical objects across modalities. Using this system, Dataset~1 (1200 frames) had 800 frames annotated for a total of 5815 identical objects, while Dataset~2 (600 frames) had 200 frames annotated for 619 identical objects.

\subsubsection{Training Details}
All deep learning models were trained from scratch on the UArizona High-Performance Computing Platform, which featured a single Nvidia 32GB V100S GPU, an AMD Zen2 processor with 5 cores, and 30\,GB of RAM. Training used PyTorch~2.0 with the Adam optimizer (momentum of 0.937, weight decay of \(5 \times 10^{-4}\)), a cosine learning rate schedule starting at 0.001 and decaying to 0.00001, and a 0.05 warm-up ratio. An exponential moving average (EMA) with a decay rate of 0.9999 was applied for added stability. Both datasets were split into training, validation, and test sets, with 90\% used for training and validation and the remaining 10\% reserved for testing. Within the training-validation split, 90\% was allocated for training and 10\% for validation, and no data augmentation was used other than resizing. The Common Feature Discriminator was trained for 300 epochs with a batch size of 4, while the Attention-based Model was trained for 800 epochs with a batch size of 8 and a token length of 256.

\begin{figure*}[]
    \centering
    \begin{subfigure}[t]{0.47\textwidth}
        \centering
        \includegraphics[width=\textwidth]{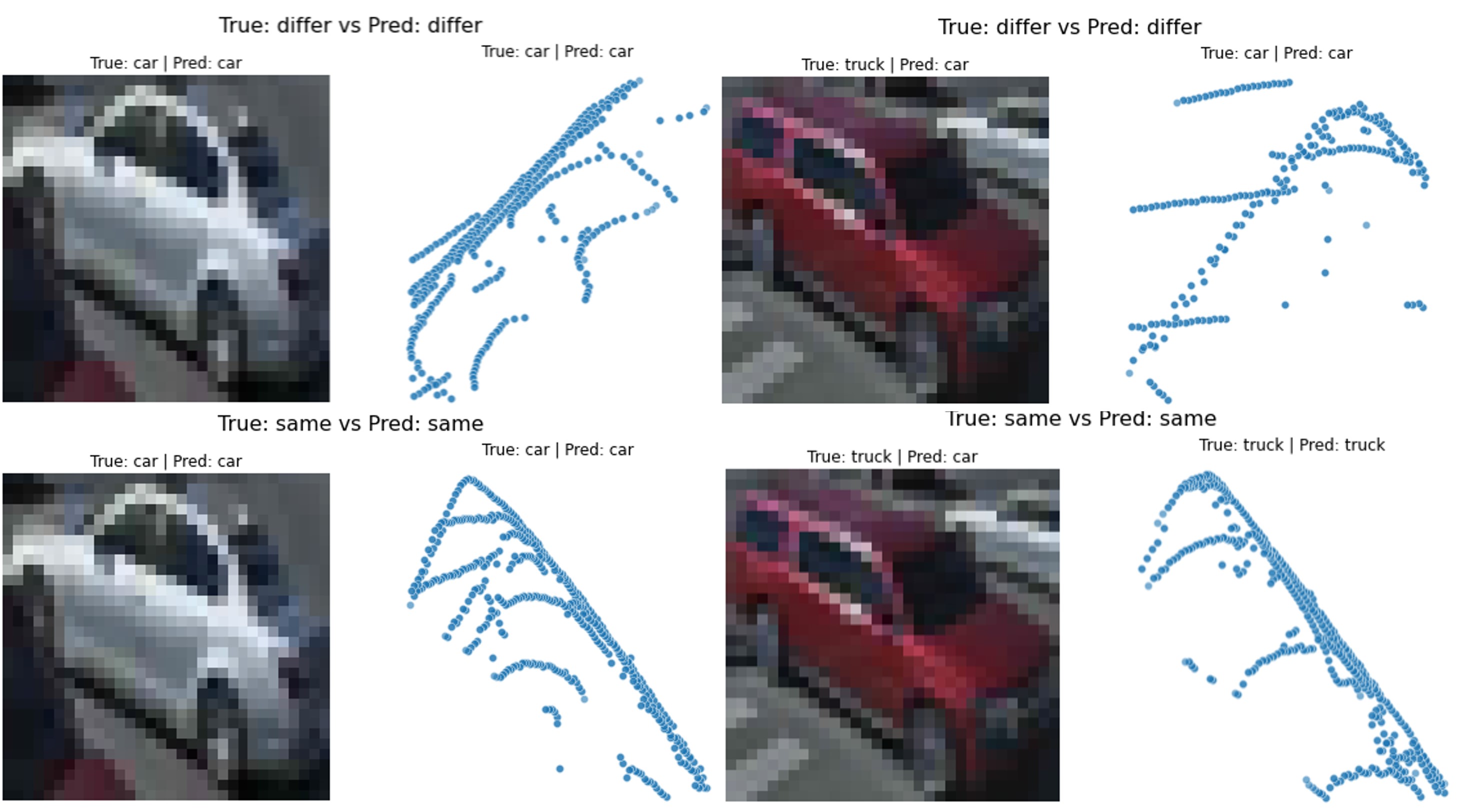}
        \caption{}
        \label{fig:1}
    \end{subfigure}\hspace{0.01\textwidth}
    \begin{subfigure}[t]{0.47\textwidth}
        \centering
        \includegraphics[width=\textwidth]{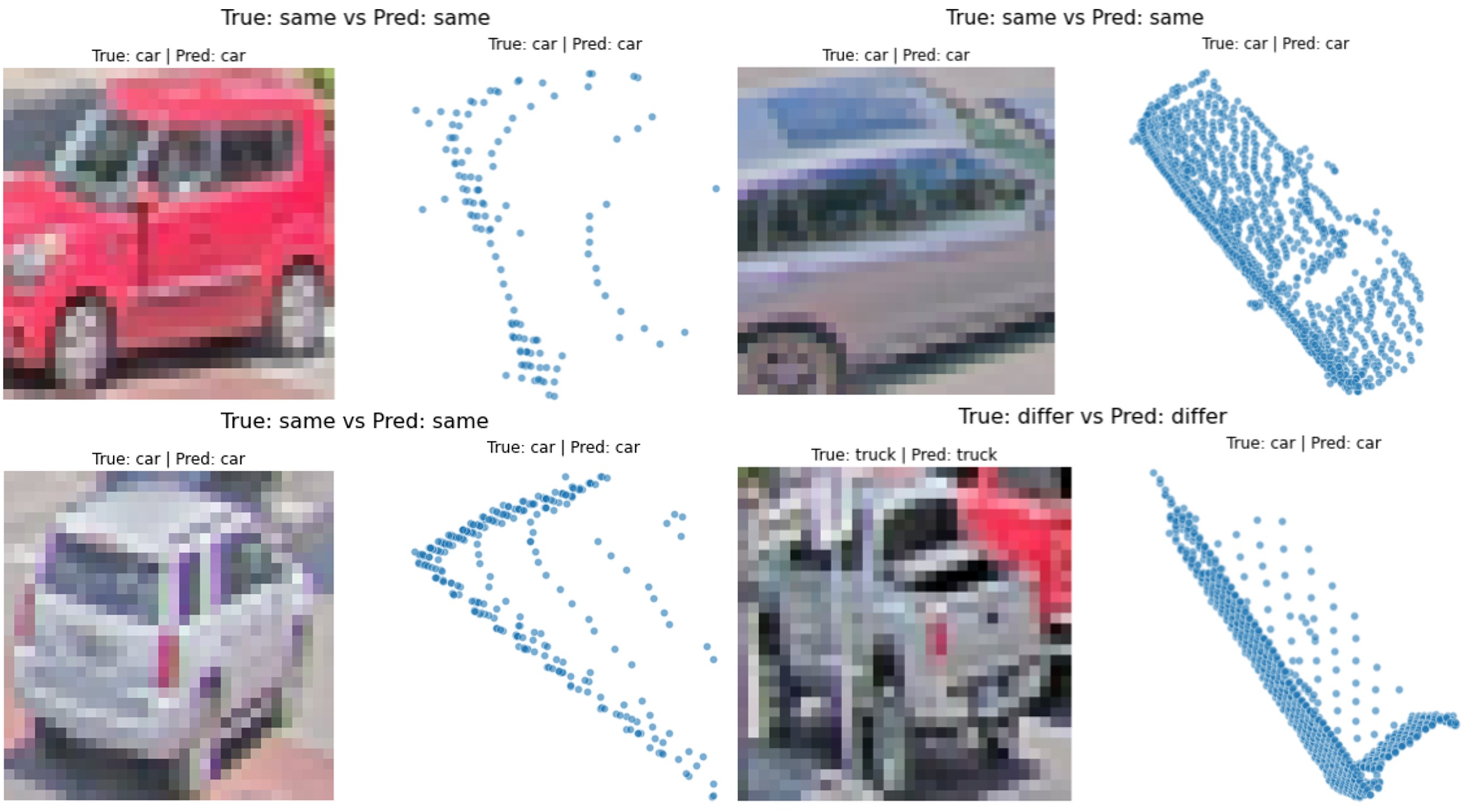}
        \caption{}
        \label{fig:2}
    \end{subfigure}
    \caption{Test Examples with Common Feature Discriminator: (a) Results on Dataset 1, (b) Results on Dataset 2.}
    \label{fig:common}
\end{figure*}

\begin{figure*}[]
    \centering
    \begin{subfigure}[t]{0.38\textwidth}
        \centering
        \includegraphics[width=\textwidth]{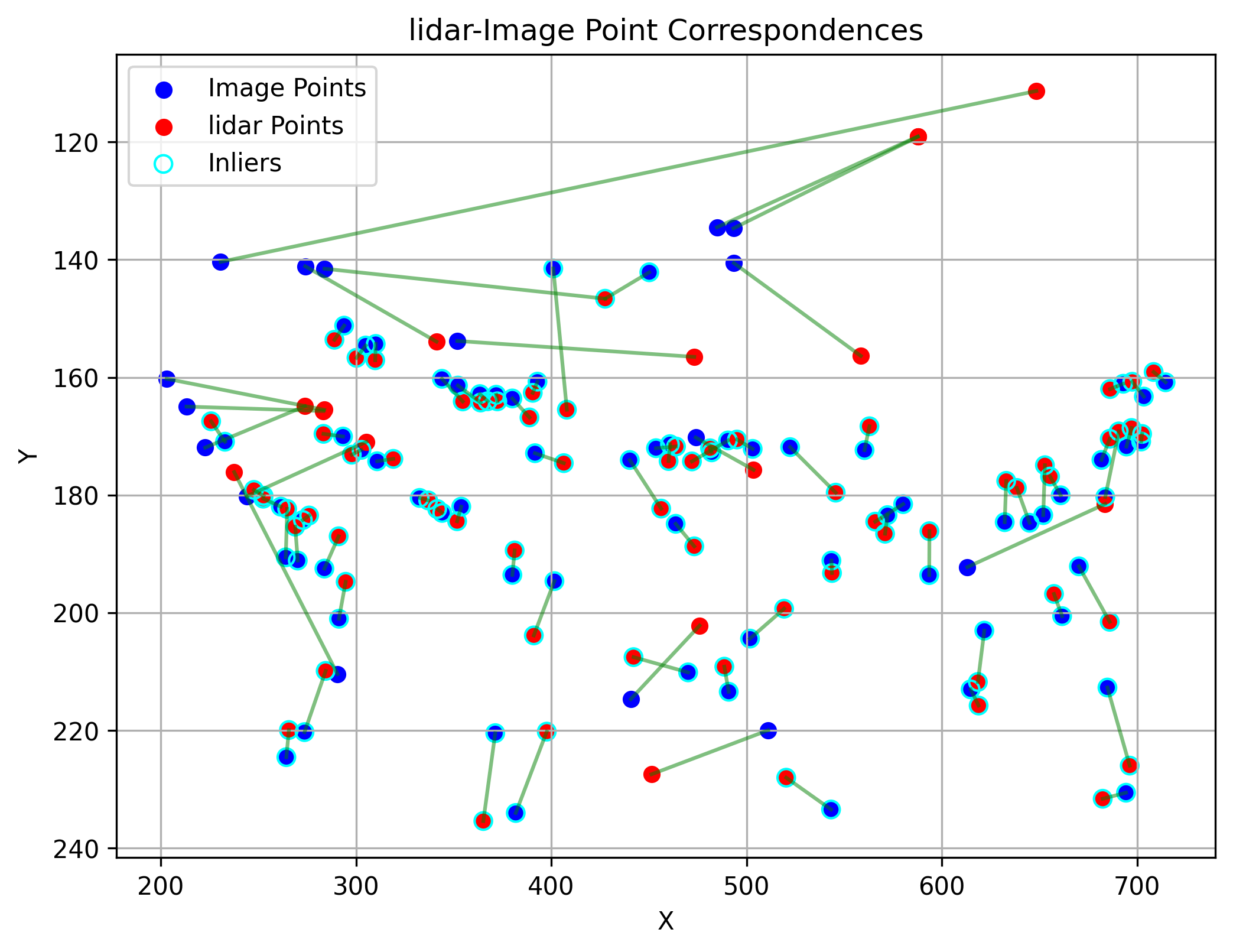}
        \caption{}
        \label{fig2:1}
    \end{subfigure}\hspace{0.06\textwidth}
    \begin{subfigure}[t]{0.45\textwidth}
        \centering
        \includegraphics[width=\textwidth]{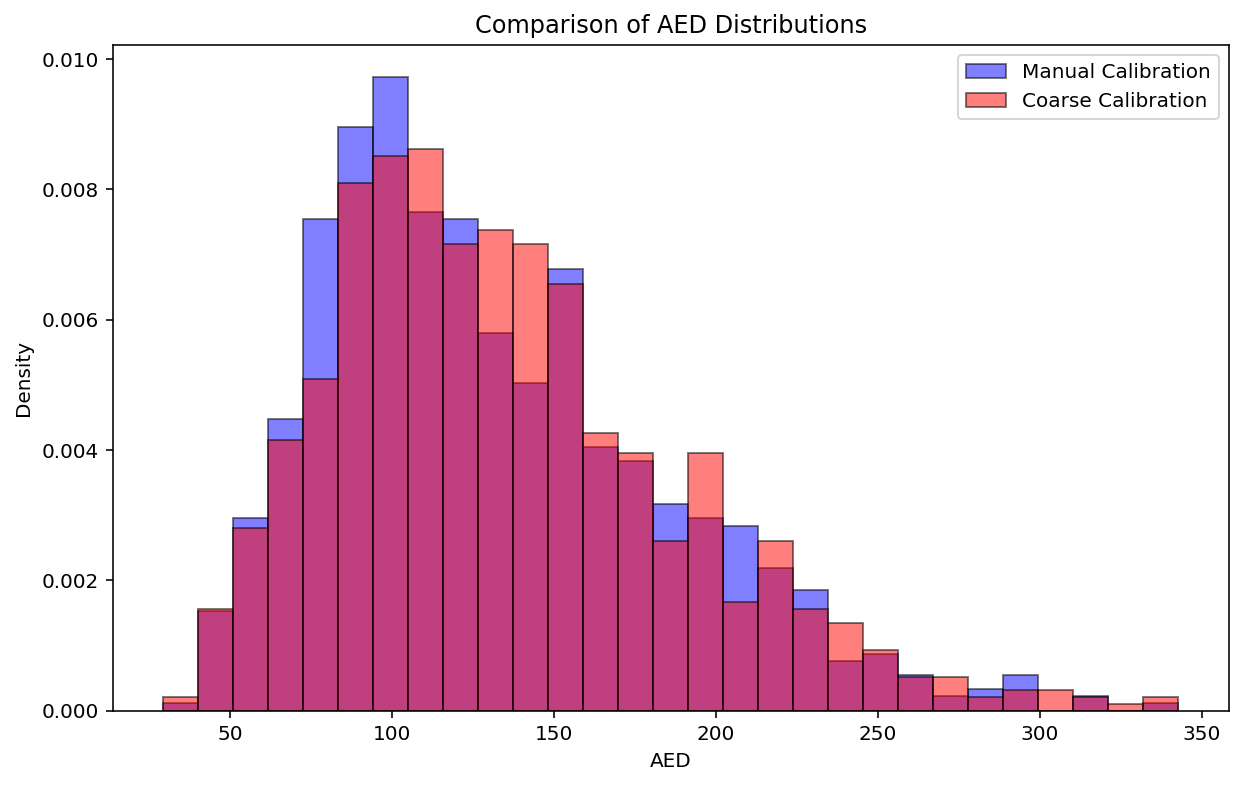}
        \caption{}
        \label{fig2:2}
    \end{subfigure}
    \caption{(a) Point Pairs Identified by the Common Feature Discriminator on Dataset 1. (b) Comparison of $\mathcal{E}_{\text{AED}}$ Distributions on Dataset 1 between Manual Calibration and Coarse Calibration.}
    \label{fig:dataset1}
\end{figure*}

\subsection{Results and Discussion}
\subsubsection{Coarse Calibration with Common Feature}
LiDAR–camera extrinsic calibration fundamentally relies on establishing point correspondences by identifying the same objects in both sensor views. Conventionally, one might manually compare camera images with LiDAR data to locate matching targets, but this process is time-consuming, labor-intensive, and error-prone—particularly given the sparse, texture-limited nature of LiDAR data compared to imagery. To address these challenges, we develop a {Common Feature Discriminator} that automatically detects and associates the same objects from both LiDAR and camera frames, thereby generating the point pairs needed for calibration.

\begin{table}[h] 
    \centering
    \caption{Performance of Common Feature Discriminator}
    \label{tab:performance}
    \begin{tabular}{lcc}
        \toprule
        \textbf{Metric} & \textbf{Dataset 1} & \textbf{Dataset 2} \\
        \midrule
        Binary Classification Accuracy (\%) & 98.00 & 92.50 \\
        Image Classification Accuracy (\%)  & 82.80 & 72.00 \\
        LiDAR Classification Accuracy (\%)  & 87.34 & 85.50 \\
        \bottomrule
    \end{tabular}
\end{table}

\paragraph{Common Feature Discriminator Performance}  
Table~\ref{tab:performance} summarizes the Common Feature Discriminator’s performance on both datasets, revealing consistently strong binary classification accuracies (98.00\% and 92.50\% for Datasets 1 and 2, respectively). These high scores indicate that the discriminator is highly effective at distinguishing whether pairs of LiDAR and camera detections correspond to the same physical object. Meanwhile, the slightly lower image and LiDAR classification accuracies for Dataset 2 reflect the inherent variability in each modality’s appearance and point cloud density, as well as the increased complexity of Dataset~2’s urban traffic scenes. Fig.~\ref{fig:common} further illustrates the model’s qualitative behavior: two distinct objects (“differ”) are correctly identified as different, while two identical objects from different sensor views are consistently classified as “same.” This underlines the model’s robustness when handling variations in object types and poses. Notably, though occasional misclassifications occur—such as trucks being predicted as cars—these errors are relatively rare and do not significantly affect the system’s ability to produce reliable point correspondences.

\begin{figure*}[h]
	\centering
	\includegraphics[width=0.95\textwidth]{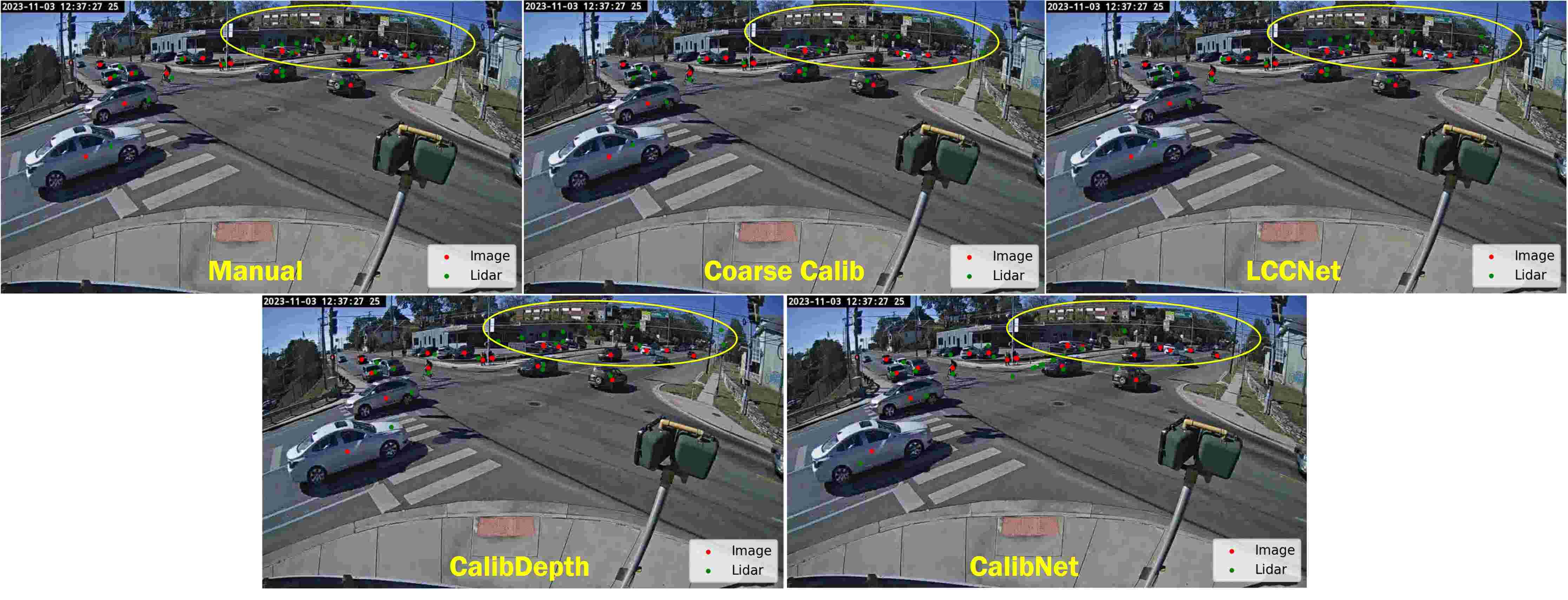}
	\caption{Example Images showing Calibration Results from Coarse Calibration and Other Methods.}
	\label{coarse_calib}
\end{figure*}

\begin{figure*}[]
	\centering
	\includegraphics[width=0.93\textwidth]{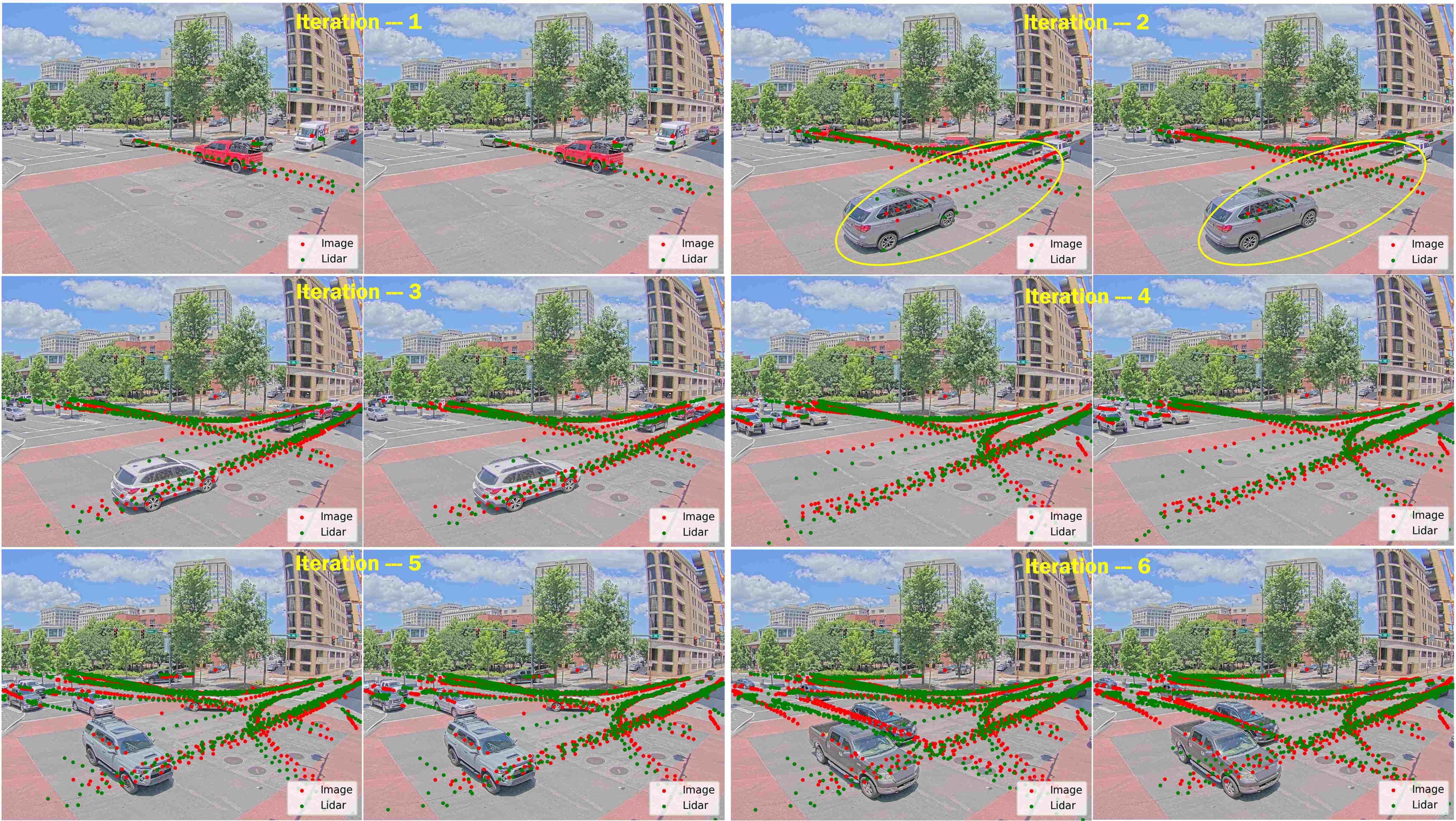}
	\caption{Trajectory Images illustrating the Calibration Performance Evolution through Iterative Refined Calibration.}
	\label{evo_imgs}
\end{figure*}

\begin{figure*}[]
    \centering
    \begin{subfigure}[t]{0.46\textwidth}
        \centering
        \includegraphics[width=\textwidth]{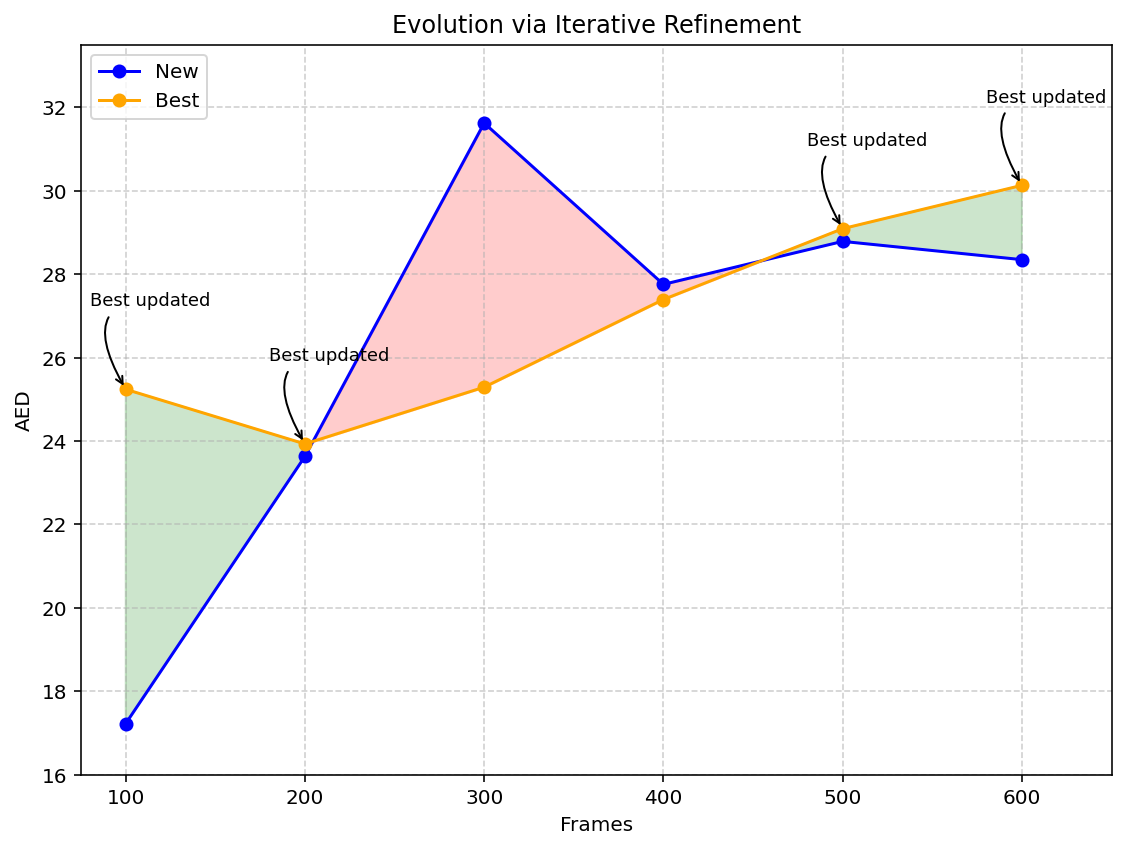}
        \caption{}
        \label{fig3:1}
    \end{subfigure}\hspace{0.001\textwidth}
    \begin{subfigure}[t]{0.46\textwidth}
        \centering
        \includegraphics[width=\textwidth]{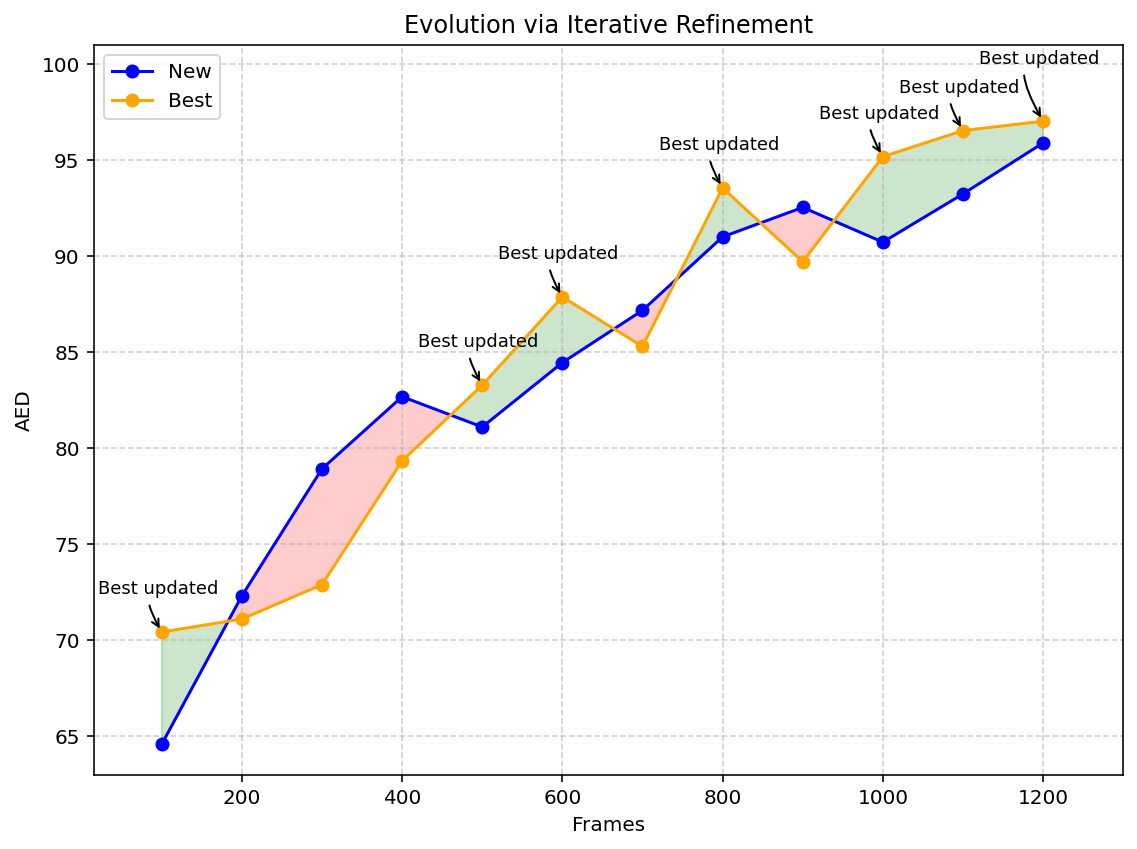}
        \caption{}
        \label{fig3:2}
    \end{subfigure}
    \caption{Evolution of $\mathcal{E}_{\text{AED}}$ over Frames during Iterative Refinement: (a) on Dataset 1, (b) on Dataset 2. Green-shaded areas indicate $\mathcal{E}_{\text{AED}}$ improvement (Best matrix updated); Red-shaded areas indicate $\mathcal{E}_{\text{AED}}$ degradation (Best matrix unchanged). }
    \label{fig:evo_plot}
\end{figure*}

\begin{table*}[]
    \centering
    \caption{Performance Comparison of Coarse Calibration and Other Methods}
    \setlength{\tabcolsep}{4pt}
    \renewcommand{\arraystretch}{1.2}
    \begin{adjustbox}{width=0.9\textwidth}
    \begin{tabular}{l cc cc cc cc cc}
        \toprule
        &
        \multicolumn{2}{c}{\textbf{Manual}} &
        \multicolumn{2}{c}{\textbf{Coarse}} &
        \multicolumn{2}{c}{\textbf{LCCNet}} &
        \multicolumn{2}{c}{\textbf{CalibDepth}} &
        \multicolumn{2}{c}{\textbf{CalibNet}} \\
        
        \cmidrule(lr){2-3}
        \cmidrule(lr){4-5}
        \cmidrule(lr){6-7}
        \cmidrule(lr){8-9}
        \cmidrule(lr){10-11}
         & $\mathcal{E}_{\text{AED}}$ & $\mathcal{E}_{\text{RMSE}}$ & $\mathcal{E}_{\text{AED}}$ & $\mathcal{E}_{\text{RMSE}}$ & $\mathcal{E}_{\text{AED}}$ & $\mathcal{E}_{\text{RMSE}}$ & $\mathcal{E}_{\text{AED}}$ & $\mathcal{E}_{\text{RMSE}}$ & $\mathcal{E}_{\text{AED}}$ & $\mathcal{E}_{\text{RMSE}}$ \\
        \midrule
        
        {Dataset 1} 
        & 131.04 & 111.57
        & 134.74 & 114.64
        & 133.55 & 115.65
        & 137.94 & 118.12
        & 140.82 & 126.91
        \\
        
        {Dataset 2} 
        & 40.79  & 32.70
        & 36.39  & 28.72
        & 29.71  & 24.31
        & 46.57  & 38.35
        & 53.22  & 45.03
        \\
        
        \bottomrule
    \end{tabular}
    \end{adjustbox}
    \label{tab:coarse_comparison}
\end{table*}

\begin{figure*}[]
    \centering
    \begin{subfigure}[t]{0.32\textwidth}
        \centering
        \includegraphics[width=\textwidth]{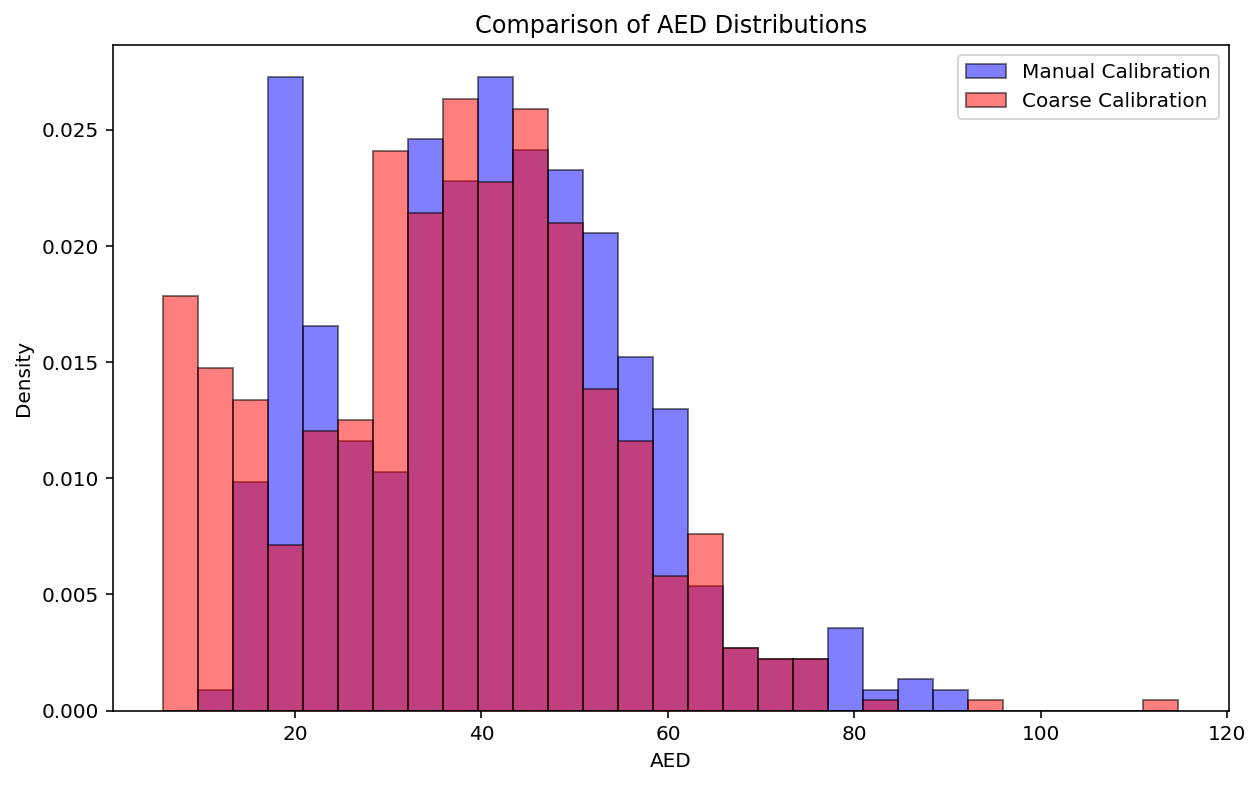}
        \caption{}
        \label{fig4:1}
    \end{subfigure}
    \begin{subfigure}[t]{0.32\textwidth}
        \centering
        \includegraphics[width=\textwidth]{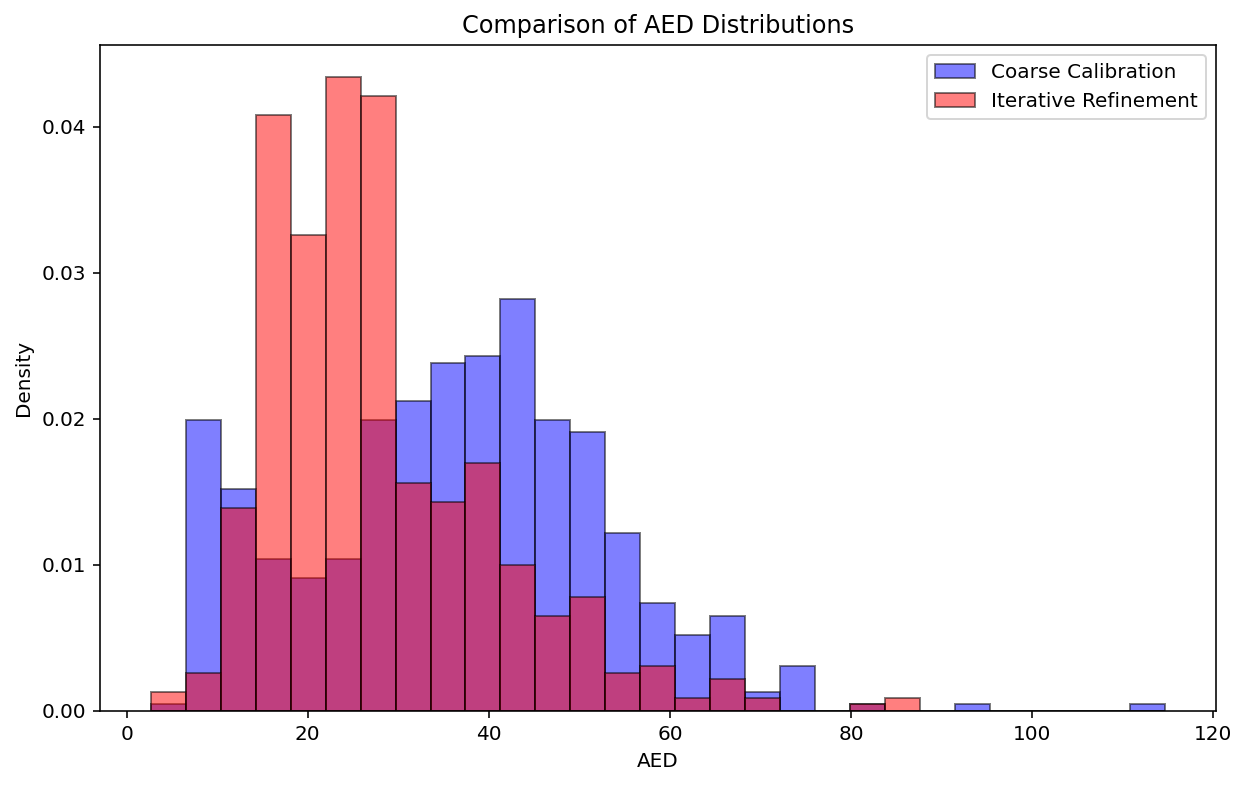}
        \caption{}
        \label{fig4:2}
    \end{subfigure}
    \begin{subfigure}[t]{0.32\textwidth}
        \centering
        \includegraphics[width=\textwidth]{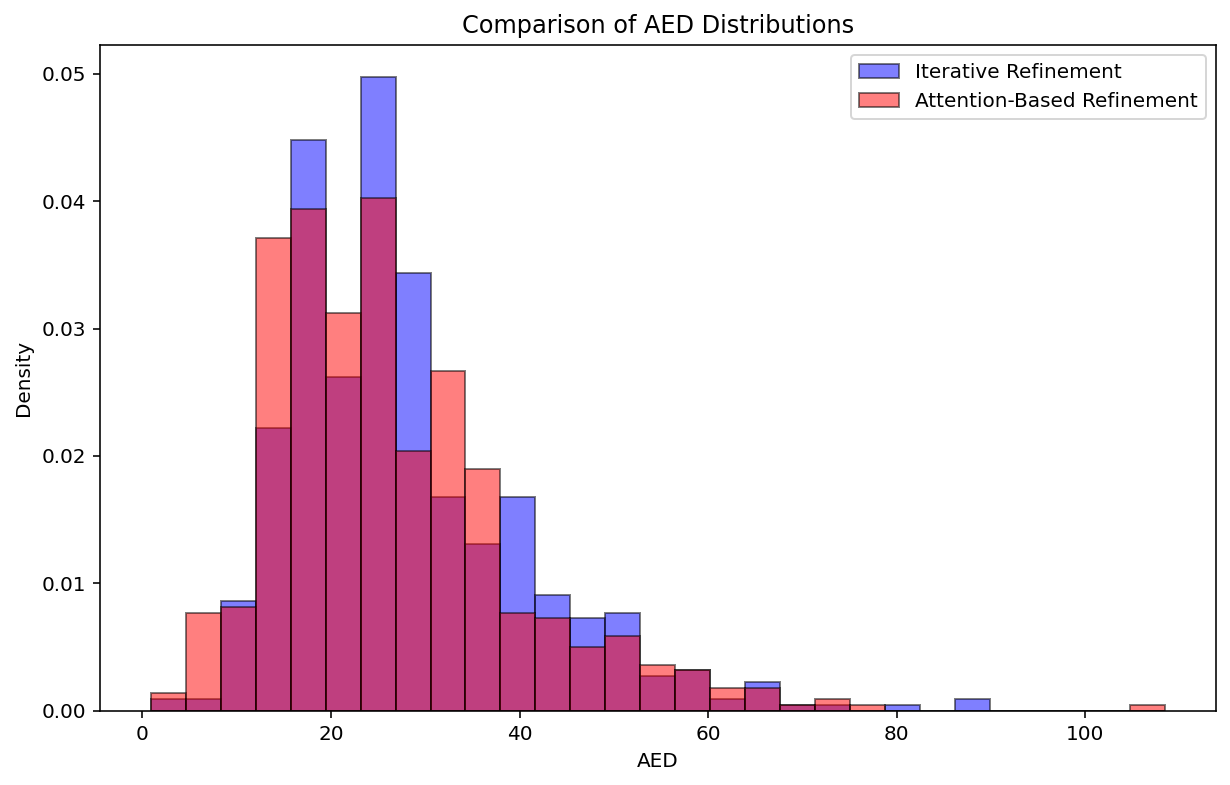}
        \caption{}
        \label{fig4:3}        
    \end{subfigure}
    \caption{Comparison of $\mathcal{E}_{\text{AED}}$ Distributions on Dataset 2 between: (a) Manual and Coarse Calibration, (b) Coarse and Iterative Refined Calibration, (c) Iterative Refined and Attention-Based Calibration. }
    \label{fig:3_hist}
\end{figure*}

\begin{figure*}[]
	\centering
	\includegraphics[width=0.95\textwidth]{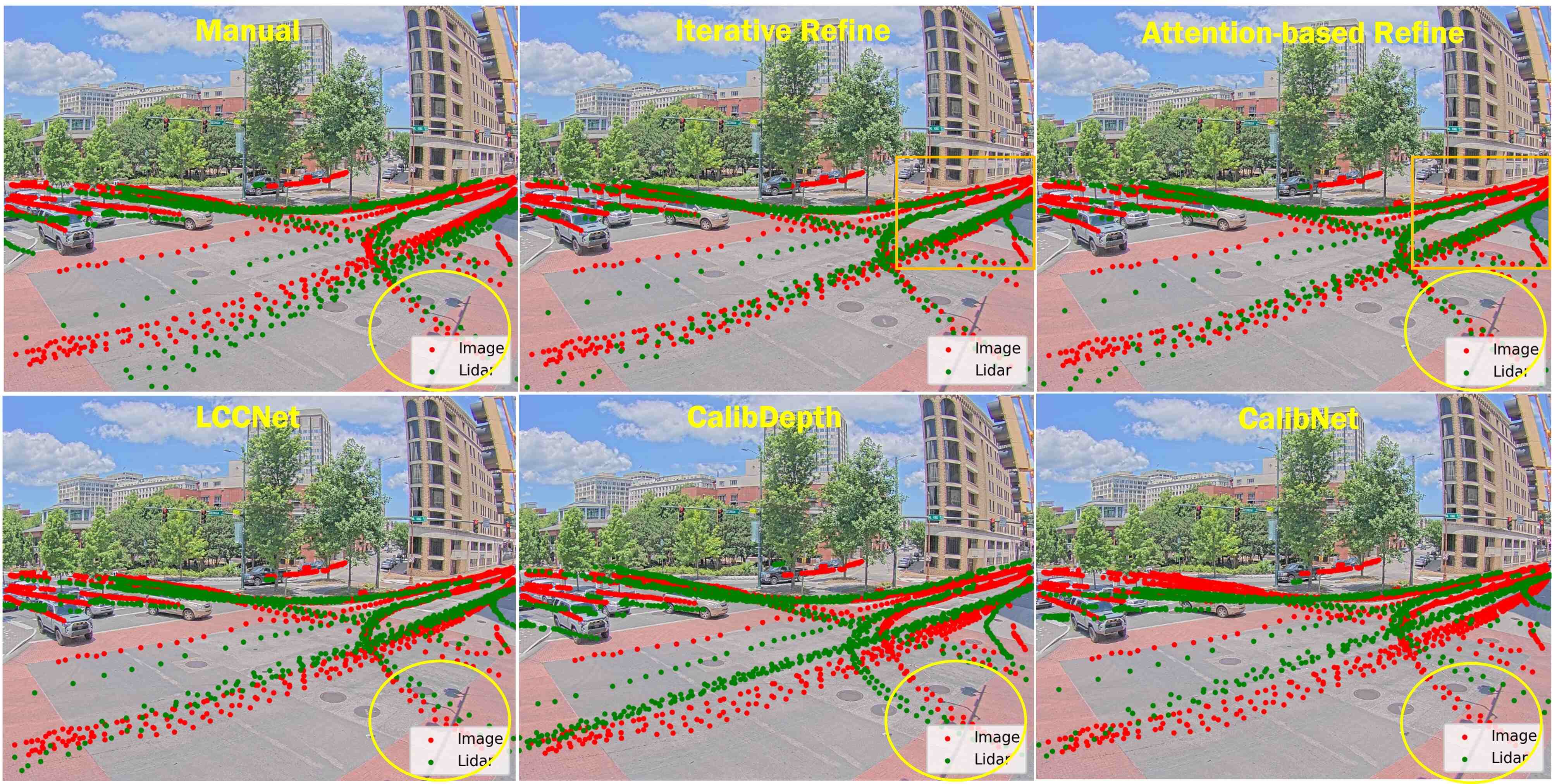}
	\caption{Trajectory Images comparing the Calibration Results after Attention-Based Refinement with Other Methods.}
	\label{atte_traj}
\end{figure*}

\paragraph{Coarse Calibration Accuracy}
Once the Common Feature Discriminator identifies matching objects across LiDAR and camera modalities and the corresponding point pairs are derived (as shown in Fig. \ref{fig2:1}), Homography calibration is employed to obtain a coarse calibration matrix. To comprehensively evaluate the accuracy of this coarse solution, we compare it with several existing calibration methods: Manual calibration, LCCNet \cite{lv2021lccnet}, CalibDepth \cite{zhu2023calibdepth}, and CalibNet \cite{8593693}. It is worth noting that for Manual calibration, we selected 34 representative point pairs uniformly distributed across the sensors' fields of view through manual object matching. Due to the time-intensive nature of this process, we did not exhaustively select all possible point pairs. Thus, the manual calibration results presented here represent a typical calibration effort within a reasonable timeframe, rather than a full-effort exhaustive manual optimization. Table \ref{tab:coarse_comparison} presents the results in terms of the reprojection error metrics \(\mathcal{E}_{\text{AED}}\) and \(\mathcal{E}_{\text{RMSE}}\) (defined in Section~\ref{homo_sec}).
From Table \ref{tab:coarse_comparison}, our coarse calibration demonstrates comparable or, in some cases, superior performance compared to other methods. Specifically, the automated coarse calibration outperforms Manual calibration on Dataset 2, although it exhibits a slightly higher reprojection error than Manual calibration on Dataset 1 (as visualized in Fig. \ref{fig2:2}). Nevertheless, the significant advantages of the automated approach in real-time operation and reduced human intervention render this trade-off both acceptable and practical. Furthermore, while the coarse method occasionally exhibits slightly higher errors than certain deep learning–based solutions (e.g., LCCNet), it consistently surpasses others (e.g., CalibDepth and CalibNet), underscoring the effectiveness of the proposed strategy. Fig. \ref{coarse_calib} presents example calibration outcomes, with red dots (camera detections) and green dots (LiDAR detections) projected onto the image plane. Despite some minor misalignments—particularly in the upper portion of the scene (highlighted by the yellow ellipse)—the coarse calibration overall provides a notably tighter alignment between the two sensor views, potentially enabling precise LiDAR-camera fusion in real-world traffic scenarios.

\begin{table}[]
    \centering
    \caption{Reprojection Error Evolution in Iterative Refinement for Dataset 1}
    \setlength{\tabcolsep}{6pt}
    \renewcommand{\arraystretch}{1.2}
    
    \begin{adjustbox}{width=0.499\textwidth}
    \begin{tabular}{cccc} 
        \toprule
        \textbf{Frame Interval} &
        \textbf{\(\mathcal{E}_{\text{AED}}\) (New)} &
        \textbf{\(\mathcal{E}_{\text{AED}}\) (Best)} &
        \textbf{Best Updated} \\
        \midrule
        0--100 & 17.2232 & 25.2391 & \textbf{Yes} (Best ← New) \\
        0--200 & 23.6301 & 23.9287 & \textbf{Yes} (Best ← New) \\
        0--300 & 31.6232 & 25.2909 & No \\
        0--400 & 27.7557 & 27.3924 & No \\
        0--500 & 28.7860 & 29.0907 & \textbf{Yes} (Best ← New) \\
        0--600 & 28.3482 & 30.1302 & \textbf{Yes} (Best ← New) \\
        \bottomrule
    \end{tabular}
    \end{adjustbox}
    \label{tab:iterative_opt}
\end{table}

\subsubsection{Fine Calibration with Iterative Refinement}
Building on the coarse calibration matrix, the iterative refinement process addresses two key objectives: (1) mitigating the imperfect object matching inherent in the coarse calibration's Common Feature Discriminator, and (2) enhancing calibration accuracy, reliability, and robustness through the iterative integration of additional point pairs into the optimization process. As outlined in Algorithm~\ref{alg:iter}, the method periodically aggregates newly formed point correspondences over successive frames to redo the Homography calibration and updates the calibration matrix whenever a lower reprojection error is achieved. 

\begin{table}[ht]
    \centering
    \caption{Reprojection Error Evolution in Iterative Refinement for Dataset 2}
    \setlength{\tabcolsep}{6pt}
    \renewcommand{\arraystretch}{1.2}
    
    \begin{adjustbox}{width=0.499\textwidth}
    \begin{tabular}{cccc} 
        \toprule
        \textbf{Frame Interval} &
        \textbf{\(\mathcal{E}_{\text{AED}}\) (New)} &
        \textbf{\(\mathcal{E}_{\text{AED}}\) (Best)} &
        \textbf{Best Updated} \\
        \midrule
        0--100  & 64.582 & 70.423 & \textbf{Yes} (Best \(\leftarrow\) New) \\
        0--200  & 72.307 & 71.101 & No \\
        0--300  & 78.922 & 72.894 & No \\
        0--400  & 82.678 & 79.334 & No \\
        0--500  & 81.099 & 83.277 & \textbf{Yes} (Best \(\leftarrow\) New) \\
        0--600  & 84.451 & 87.872 & \textbf{Yes} (Best \(\leftarrow\) New) \\
        0--700  & 87.173 & 85.293 & No \\
        0--800  & 90.998 & 93.546 & \textbf{Yes} (Best \(\leftarrow\) New) \\
        0--900  & 92.534 & 89.708 & No \\
        0--1000 & 90.724 & 95.177 & \textbf{Yes} (Best \(\leftarrow\) New) \\
        0--1100 & 93.234 & 96.532 & \textbf{Yes} (Best \(\leftarrow\) New) \\
        0--1200 & 95.891 & 97.023 & \textbf{Yes} (Best \(\leftarrow\) New) \\
        \bottomrule
    \end{tabular}
    \end{adjustbox}
    \label{tab:iterative_opt_2}
\end{table}

Tables~\ref{tab:iterative_opt} and~\ref{tab:iterative_opt_2} detail the reprojection error evolution (using the \(\mathcal{E}_{\text{AED}}\) metric) at different frame intervals (with an interval of 100 frames in our implementation) for Datasets~1 and~2. In each interval, the algorithm determines whether the newly computed homography matrix (\textit{New}) provides a tighter alignment than the previously best-known matrix (\textit{Best}); if so, it updates the calibration accordingly. Fig.~\ref{fig3:1} and~\ref{fig3:2} visualize these updates, where the blue line denotes the error obtained from the newly recalibrated matrix in each iteration, and the orange line tracks the evolving best-known solution. Not every recalibration step yields an improvement—reflecting the inherent noise and variability of real-world data—but key frame intervals (e.g., 0--100 for Dataset 1 and 0--1000 for Dataset 2) demonstrate significant error reductions, confirming that the iterative approach converges toward a more accurate solution over time. These updates demonstrate the iterative optimization process's ability to adaptively refine the calibration as additional data and correspondences become available, ultimately enabling the iterative refinement to achieve significantly higher accuracy compared to the initial coarse calibration (as shown in Fig. \ref{fig4:2}).

\begin{table*}[]
    \centering
    \caption{Performance Comparison of Iterative Refined Calibration and Other Methods}
    \setlength{\tabcolsep}{4pt}
    \renewcommand{\arraystretch}{1.2}
    \begin{adjustbox}{width=0.9\textwidth}
    \begin{tabular}{l cc cc cc cc cc}
        \toprule
        &
        \multicolumn{2}{c}{\textbf{Manual}} &
        \multicolumn{2}{c}{\textbf{Iterative}} &
        \multicolumn{2}{c}{\textbf{LCCNet}} &
        \multicolumn{2}{c}{\textbf{CalibDepth}} &
        \multicolumn{2}{c}{\textbf{CalibNet}} \\
        
        \cmidrule(lr){2-3}
        \cmidrule(lr){4-5}
        \cmidrule(lr){6-7}
        \cmidrule(lr){8-9}
        \cmidrule(lr){10-11}
         & $\mathcal{E}_{\text{AED}}$ & $\mathcal{E}_{\text{RMSE}}$ & $\mathcal{E}_{\text{AED}}$ & $\mathcal{E}_{\text{RMSE}}$ & $\mathcal{E}_{\text{AED}}$ & $\mathcal{E}_{\text{RMSE}}$ & $\mathcal{E}_{\text{AED}}$ & $\mathcal{E}_{\text{RMSE}}$ & $\mathcal{E}_{\text{AED}}$ & $\mathcal{E}_{\text{RMSE}}$ \\
        \midrule
        
        {Dataset 1} 
        & 131.04 & 111.57
        & 95.89 & 74.10
        & 133.55 & 115.65
        & 137.94 & 118.12
        & 140.82 & 126.91
        \\
        
        {Dataset 2} 
        & 40.79  & 32.70
        & 28.35  & 23.09
        & 29.71  & 24.31
        & 46.57  & 38.35
        & 53.22  & 45.03
        \\
        
        \bottomrule
    \end{tabular}
    \end{adjustbox}
    \label{tab:iter_comparison}
\end{table*}

\begin{table*}[]
    \centering
    \caption{Performance Comparison of CalibRefine and Other Methods}
    \setlength{\tabcolsep}{4pt}
    \renewcommand{\arraystretch}{1.2}
    \begin{adjustbox}{width=0.9\textwidth}
    \begin{tabular}{l cc cc cc cc cc}
        \toprule
        &
        \multicolumn{2}{c}{\textbf{Manual}} &
        \multicolumn{2}{c}{\textbf{CalibRefine}} &
        \multicolumn{2}{c}{\textbf{LCCNet}} &
        \multicolumn{2}{c}{\textbf{CalibDepth}} &
        \multicolumn{2}{c}{\textbf{CalibNet}} \\
        
        \cmidrule(lr){2-3}
        \cmidrule(lr){4-5}
        \cmidrule(lr){6-7}
        \cmidrule(lr){8-9}
        \cmidrule(lr){10-11}
         & $\mathcal{E}_{\text{AED}}$ & $\mathcal{E}_{\text{RMSE}}$ & $\mathcal{E}_{\text{AED}}$ & $\mathcal{E}_{\text{RMSE}}$ & $\mathcal{E}_{\text{AED}}$ & $\mathcal{E}_{\text{RMSE}}$ & $\mathcal{E}_{\text{AED}}$ & $\mathcal{E}_{\text{RMSE}}$ & $\mathcal{E}_{\text{AED}}$ & $\mathcal{E}_{\text{RMSE}}$ \\
        \midrule
        
        {Dataset 1} 
        & 131.04 & 111.57
        & 93.27 & 72.68
        & 133.55 & 115.65
        & 137.94 & 118.12
        & 140.82 & 126.91
        \\
        
        {Dataset 2} 
        & 40.79  & 32.70
        & 26.40  & 22.25
        & 29.71  & 24.31
        & 46.57  & 38.35
        & 53.22  & 45.03
        \\
        
        \bottomrule
    \end{tabular}
    \end{adjustbox}
    \label{tab:final_comparison}
\end{table*}

\begin{figure}[]
	\centering
	\includegraphics[width=0.49\textwidth]{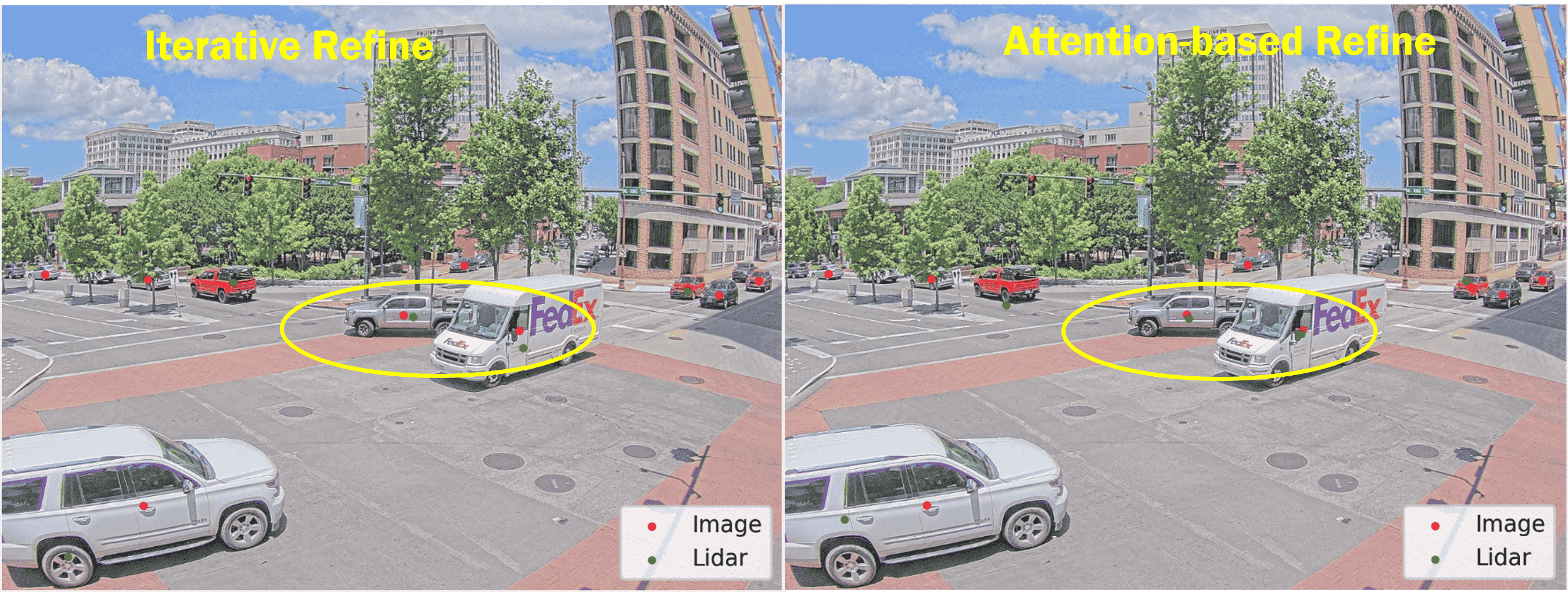}
	\caption{Improved results with Attention-based Refinement over Iterative Refinement.}
	\label{atte_iter}
\end{figure}

\begin{figure*}[h]
	\centering
	\includegraphics[width=0.95\textwidth]{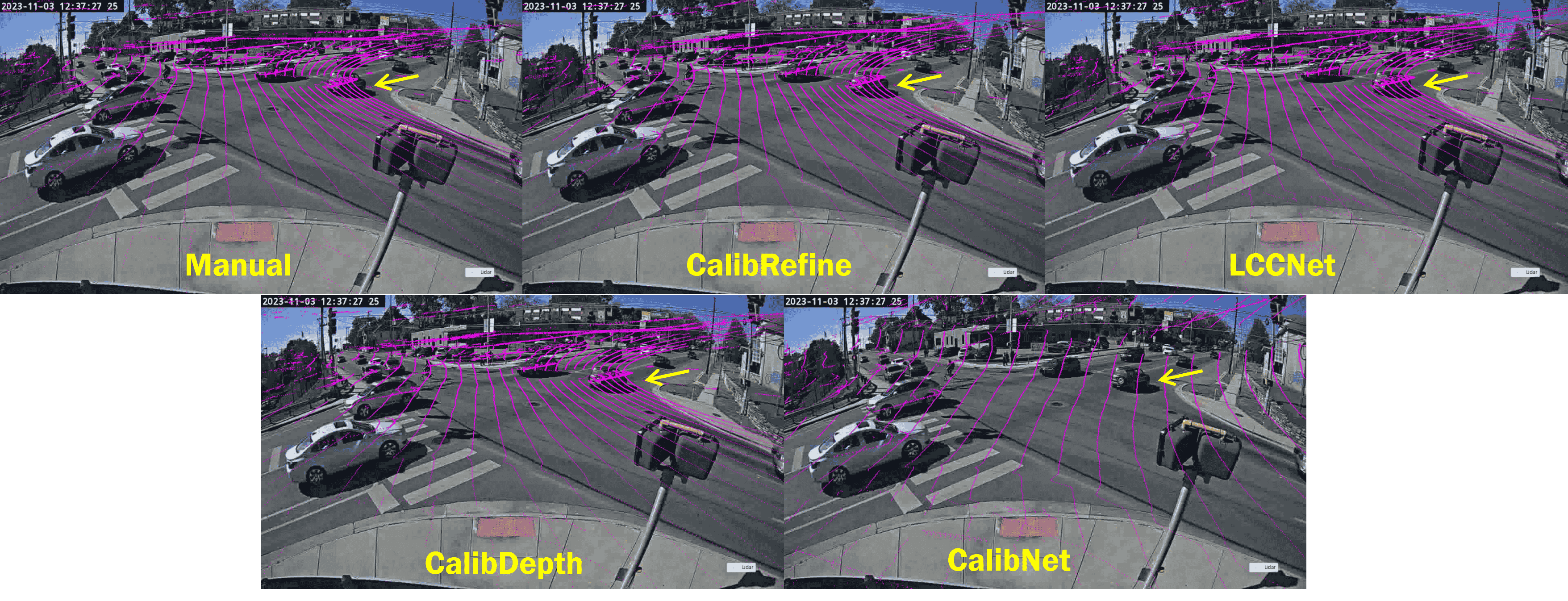}
	\caption{Comparison of LiDAR Point Cloud Projection Results on Dataset 1 using Different Calibration Methods.}
	\label{PCD_1}
\end{figure*}

Fig.~\ref{evo_imgs} provides a more detailed view of how the iterative refinement process unfolds over six iterations, as LiDAR point trajectories are progressively better aligned with camera detections. In {Iteration-1}, noticeable offsets appear in the vehicle on the left side and for several distant cars near the center of the scene, indicating that the initial coarse calibration matrix is not sufficiently accurate for all regions. By {Iteration-2}, however, there is a conspicuous improvement: the LiDAR points more precisely cluster around the corresponding vehicles—particularly the trajectory highlighted by the yellow ellipse—demonstrating that additional correspondences acquired in this step already correct many of the early misalignments. Over {Iteration-3} and {-4}, the algorithm refines the alignment further, as the expanded pool of object correspondences helps correct lingering calibration errors, especially for vehicles at varying distances. Finally, by {Iteration-5} and {-6}, the calibration has converged to a state where the majority of LiDAR returns closely coincide with the camera detections, indicating that additional correspondences spanning a broader field of view substantially improve calibration fidelity.

Table~\ref{tab:iter_comparison} compares the final calibration performance with the aforementioned methods. Notably, the iterative refinement outperforms manual calibration by a sizeable margin in both datasets, reducing \(\mathcal{E}_{\text{AED}}\) from 131.04 to 95.89 in Dataset~1 and from 40.79 to 28.35 in Dataset~2. It also consistently surpasses CalibDepth and CalibNet, while maintaining a competitive edge against LCCNet. These results demonstrate the effectiveness of iteratively incorporating new point correspondences in mitigating decalibrations and refining the sensor alignment. 

In practice, the iterative refinement process exhibits several key strengths:
1) Consistent Refinement: The reprojection error generally decreases over time, indicating effective optimization.
2) Adaptability: The process dynamically updates the calibration matrix when new correspondences improve accuracy, as seen in multiple intervals.
3) Robustness: Even during intervals where no improvement occurs, the process maintains a stable calibration without overfitting to potentially noisy correspondences.
These findings highlight the iterative refinement’s ability to achieve high-precision calibration—especially in scenarios with sufficient frame data and reliable correspondences—while ensuring continuous accuracy improvement as more data becomes available, making it a robust solution for real-world applications. 

\subsubsection{Fine Calibration with Attention-based Refinement}
Although the iterative refinement approach already demonstrates strong performance, it remains inherently limited by the planar assumptions of Homography. Our proposed attention-based refinement aims to mitigate errors caused by image distortions and non-planar surfaces. As shown in Table~\ref{tab:final_comparison}, calibration after applying attention-based refinement (i.e., \emph{CalibRefine}) achieves lower reprojection errors than other methods on both datasets, surpassing the iterative refinement (Table~\ref{tab:iter_comparison}) in most metrics. Fig.~\ref{fig4:3} offers a more granular view of these improvements by comparing the \(\mathcal{E}_{\text{AED}}\) distributions of iterative refinement and attention-based refinement. While the latter still exhibits some overlap with the former, its overall distribution skews toward smaller errors, indicating a more consistently accurate alignment between LiDAR and camera data. Fig. \ref{atte_iter} visually illustrates such performance gains of attention-based refinement over iterative refinement.

\begin{figure*}[]
	\centering
	\includegraphics[width=0.95\textwidth]{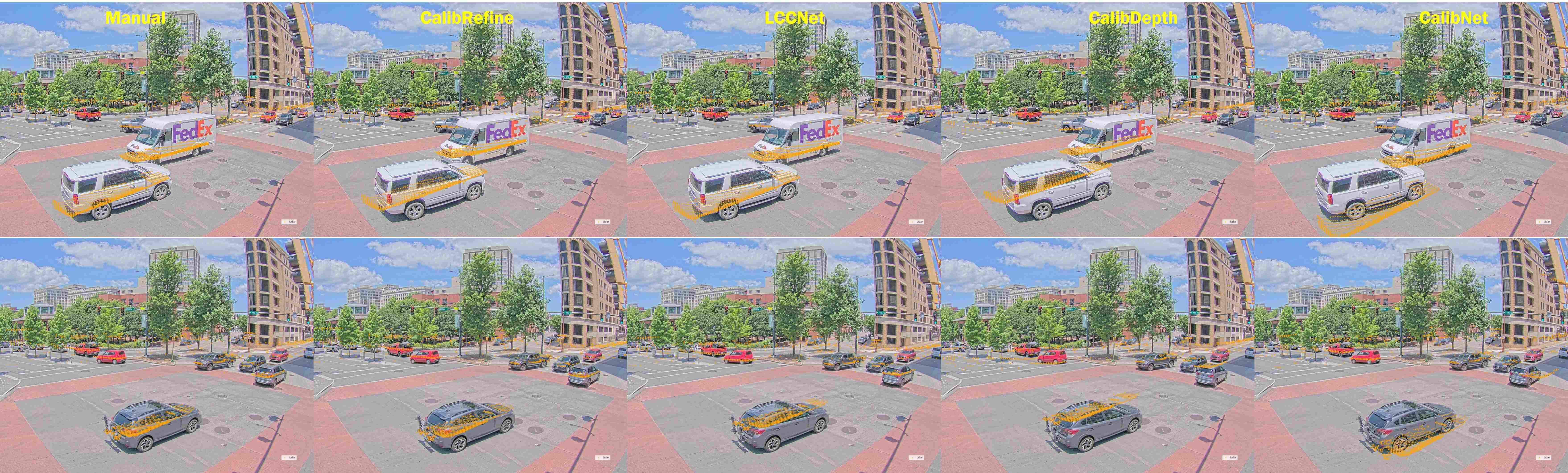}
	\caption{Comparison of LiDAR Point Cloud Projection Results on Dataset 2 using Different Calibration Methods.}
	\label{PCD_2}
\end{figure*}

\begin{figure*}[]
	\centering
	\includegraphics[width=0.95\textwidth]{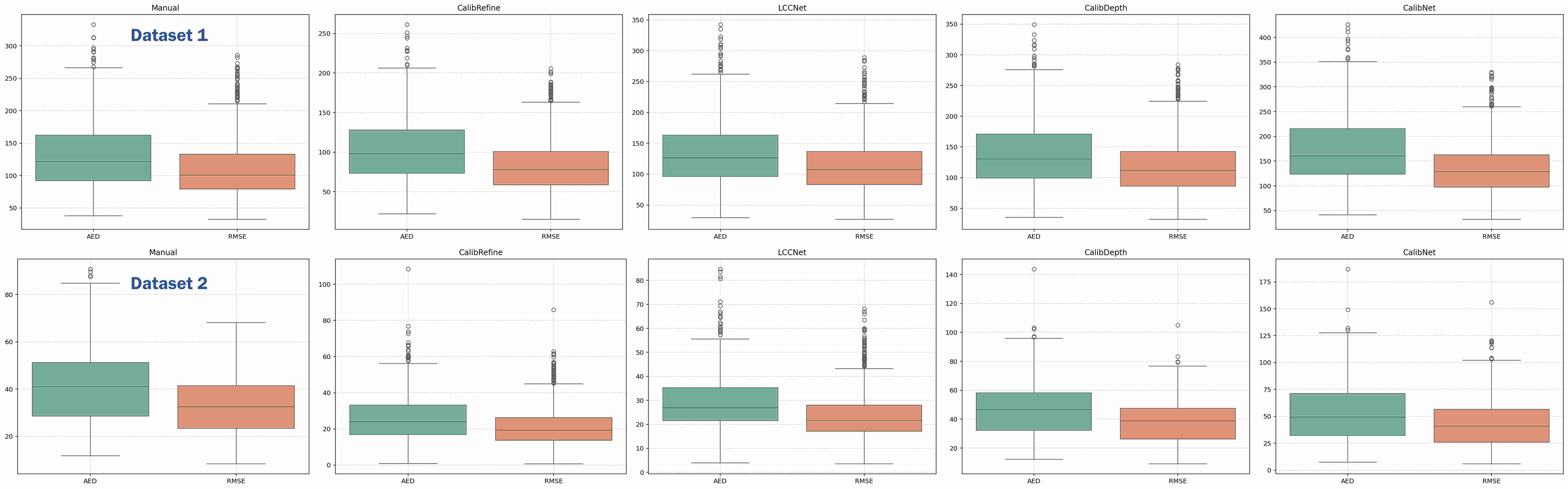}
	\caption{Calibration Error Distributions across Different Methods on Dataset 1 (top row) and Dataset 2 (bottom row).}
	\label{cal_err}
\end{figure*}

Fig.~\ref{atte_traj} further demonstrates the enhancement achieved by Attention-based Refinement compared to other methods in real-world traffic scenes. A closer examination of regions near scene edges and sidewalk corners (highlighted by orange rectangles and yellow circles) reveals that iterative refinement and purely manual alignment often exhibit limitations in accurately aligning distant objects and scene edges. In contrast, Attention-based Refinement more effectively associates LiDAR points with their corresponding objects, particularly under challenging perspective angles. While LCCNet also delivers strong performance, minor misalignments remain visible near scene edges. CalibDepth and CalibNet, however, show even poorer alignment accuracy in these regions.
Evidently, the Attention-based improvement margin over iterative refinement is relatively modest, likely due in part to the already high baseline accuracy afforded by iterative methods. Another contributing factor is the inherent limitation of a 9-parameter homography matrix in capturing the full complexity of perspective transformations. These observations highlight both the promise and limitations of the proposed method. More advanced deep learning architectures or more sophisticated mapping mechanisms could better address complex real-world distortions and further improve LiDAR–camera alignment.

\newcommand{\cmark}{\ding{51}} 
\newcommand{\xmark}{\ding{55}} 

\begin{table*}[t]
    \centering
    \caption{Ablation of CalibRefine Components on Two Datasets (\cmark: enabled, \xmark: disabled)}
    \setlength{\tabcolsep}{4pt}
    \renewcommand{\arraystretch}{1.2}
    \begin{adjustbox}{width=0.96\textwidth}
    \begin{tabular}{l c c c c c c c cc cc}
        \toprule
        \multirow{2}{*}{\textbf{Settings}} &
        \multicolumn{7}{c}{\textbf{Modules}} &
        \multicolumn{2}{c}{\textbf{Dataset 1}} &
        \multicolumn{2}{c}{\textbf{Dataset 2}} \\
        \cmidrule(lr){2-8}\cmidrule(lr){9-10}\cmidrule(lr){11-12}
        & {CFD} & {RANSAC} & {BBSample} & {GBMatch} & {Iterative} & {ViT} & {Cross-Attn}
        & \(\mathcal{E}_{\text{AED}}\!\downarrow\) & \(\mathcal{E}_{\text{RMSE}}\!\downarrow\)
        & \(\mathcal{E}_{\text{AED}}\!\downarrow\) & \(\mathcal{E}_{\text{RMSE}}\!\downarrow\) \\
        \midrule
        {Coarse}
        & \cmark & \cmark & \cmark & \xmark & \xmark & \xmark & \xmark
        & 134.74 & 114.64 & 36.39 & 28.72 \\
        {\ + Iterative}
        & \cmark & \cmark & \cmark & \cmark & \cmark & \xmark & \xmark
        & 95.89 & 74.10 & 28.35 & 23.09 \\
        {\ + Attention}
        & \cmark & \cmark & \cmark & \cmark & \cmark & \cmark & \cmark
        & 93.27 & 72.68 & 26.40 & 22.25 \\
        \bottomrule
    \end{tabular}
    \end{adjustbox}
    \label{tab:ablation_components}
\end{table*}

\begin{table*}[h!]
    \centering
    \caption{Computational Demand Analysis (per 1080p frame). GPU usage measured on an Nvidia 32GB V100S.}
    \setlength{\tabcolsep}{6pt}
    \renewcommand{\arraystretch}{1.2}
    \begin{adjustbox}{width=0.95\textwidth}
    \begin{tabular}{l l c c l}
        \toprule
        \textbf{Stage} & \textbf{Operation} & \textbf{Latency (ms)} & \textbf{GPU (\%)} & \textbf{Notes} \\
        \midrule
        CFD inference & Image det. + LiDAR clustering & \textbf{153.0} & \textbf{74} & Frame-wise  \\
        Projection \& matching & LiDAR $\!\to$ image, {GBMatch}, {BBSample} & 18.1 & 18 & Scales with \#points \\
        Iterative refinement & Recalibration on accumulated pairs& 445.0/N=4.45 & 12 & Periodic; $N{=}100$\\
        Attention refinement & ViT + Cross-Attn $\to H^\Delta$ & \textbf{166.0} & \textbf{82} & Frame-wise  \\
        \midrule
        \multicolumn{2}{r}{\textbf{Overall}} & \multicolumn{1}{c}{\textbf{FPS: $\frac{1000}{339.5}\approx\mathbf{2.95}$}} & \multicolumn{1}{c}{\textbf{GPU: 82\%}} & End-to-End \\
        \bottomrule
    \end{tabular}
    \end{adjustbox}
    \label{tab:runtime_profile}
\end{table*}

Overall, our proposed CalibRefine framework consolidates three core components—Coarse Calibration, Iterative Refinement, and Attention-Based Refinement—into a unified solution. As illustrated in Fig. \ref{fig:3_hist}, each stage progressively refines the LiDAR–camera alignment, mitigating errors introduced by imperfect correspondence matching (coarse stage), limited point redundancy (iterative stage), or planar homography assumptions (attention-based stage). 
Table~\ref{tab:final_comparison} further demonstrates that CalibRefine surpasses existing state-of-the-art methods in terms of quantitative reprojection accuracy. Beyond numerical metrics, Fig.~\ref{PCD_1} and~\ref{PCD_2} offer visual validation on Datasets~1 and~2, respectively, revealing how CalibRefine more reliably overlays LiDAR points with their corresponding image objects—particularly at scene edges and larger distances. In addition, Fig.~\ref{cal_err} examines the distribution of calibration errors (\(\mathcal{E}_{\text{AED}}\) and \(\mathcal{E}_{\text{RMSE}}\)) across competing approaches. Not only does CalibRefine exhibit a lower median error, but the overall spread of high-error outliers is also reduced, indicating its consistent performance. These findings underscore the robustness and adaptability of CalibRefine in real-world traffic environments.

\subsubsection{Ablation Study and Computation Evaluation}
Table~\ref{tab:ablation_components} quantifies the incremental effects of each stage. Moving from the coarse CFD-based baseline to the iterative refinement yields the dominant gains: on \emph{Dataset~1}, \(\mathcal{E}_{\text{AED}}\) drops by \(28.8\%\) (134.74\(\rightarrow\)95.89) and \(\mathcal{E}_{\text{RMSE}}\) by \(35.4\%\) (114.64\(\rightarrow\)74.10); on \emph{Dataset~2}, the reductions are \(22.1\%\) (36.39\(\rightarrow\)28.35) and \(19.6\%\) (28.72\(\rightarrow\)23.09), respectively. Adding the attention-based refinement provides consistent, smaller improvements beyond the iterative stage—\(\mathcal{E}_{\text{AED}}\) improves by \(2.7\%\) and \(6.9\%\), while \(\mathcal{E}_{\text{RMSE}}\) improves by \(1.9\%\) and \(3.6\%\) on \emph{Datasets 1/2}. These results indicate that coarse calibration with iterative refinement accounts for most of the accuracy gains, whereas attention offers a marginal correction for residual non-planar effects.

Table~\ref{tab:runtime_profile} reports the per-frame computational profile at 1080p on an Nvidia 32\,GB V100S (batch size 1). The end-to-end throughput is \(\approx 2.95\) FPS with peak GPU utilization \(\approx 82\%\). The runtime is dominated by \emph{Attention refinement} (166.0\,ms) and \emph{CFD inference} (153.0\,ms); \emph{Projection \& matching} is modest (18.1\,ms) but grows with the number of pairs, and the amortized cost of \emph{Iterative refinement} is small (about \(4.45\) ms/frame with \(N{=}100\)). In practice, scheduling the attention module periodically (or on detected distributional shifts) can improve effective FPS with limited accuracy loss, given the narrow gap between “+Iterative” and “+Attention” in Table~\ref{tab:ablation_components}.

\subsubsection{Limitations}
CalibRefine relies on object-level correspondences to bootstrap and refine calibration; therefore, it assumes the presence of foreground objects within the overlapping fields of view. As such, scenes lacking objects (e.g., empty roads) are not applicable and were not included in our evaluation. Our experiments focus on planar urban intersections with fixed, pole-mounted sensors, implicitly leveraging ground coplanarity through a 2D homography. Formal error bounds under non-planar conditions (e.g., ramps or multi-level structures) are not provided at this stage and will be explored in future work. The experimental scope is further limited to sensor vendor-recommended, highest-resolution settings and default deployment viewpoints; we do not sweep sensor parameters (e.g., LiDAR line count, camera resolution) or alternative placements. Although the pipeline is fully automatic and requires no human intervention, its current throughput is constrained (\(\approx 2.95\) FPS). We anticipate that real-time performance can be improved through stronger hardware, an optimized C++/TensorRT implementation, or the lighter backbones.

\section{Conclusion}
In this paper, we presented CalibRefine, an end-to-end, fully automatic, targetless, and online LiDAR–camera calibration framework that integrates three core steps—coarse calibration, iterative refinement, and attention-based refinement—into a unified pipeline. By combining robust object detection with a Common Feature Discriminator, our method circumvents the need for manually placed fiducials or human-labeled sensor parameters. The coarse calibration phase provides a strong initial alignment, which the iterative refinement then continuously improves by leveraging newly acquired point correspondences across frames. Finally, the attention-based stage applies a Vision Transformer and cross-attention to handle non-planar distortions and subtle mismatches beyond the scope of homography.
Experiments on real-world urban datasets confirm that CalibRefine achieves accurate sensor alignment comparable to, and often better than, existing methods. Moving forward, the approach could benefit from exploring more advanced deep learning architectures or sophisticated mapping mechanisms, as well as extending the attention mechanism to incorporate scene geometry. Such enhancements could enable even more precise and high-fidelity calibration, particularly in large-scale deployment scenarios.


\vspace{0.5cm}

\section*{Acknowledgments}
This work was supported by The Federal Highway Administration (FHWA) Exploratory Advanced Research (EAR) Program. Award No.: 693JJ32350028.

\bibliographystyle{IEEEtran}
{\small
\bibliography{references.bib}
}


\vspace{-25pt}

\vfill

\end{document}